\pdfoutput=1

\documentclass[runningheads]{llncs}

\usepackage{eccv}

\usepackage{eccvabbrv}

\usepackage{graphicx}
\usepackage{booktabs}
\usepackage[accsupp]{axessibility} 

\usepackage{paralist}
\usepackage{tabularx}
\usepackage[dvipsnames]{xcolor}
\usepackage{colortbl}
\usepackage{multirow}
\usepackage{amsmath}
\usepackage{titletoc}

\makeatletter
\renewcommand{\paragraph}{%
  \@startsection{paragraph}{4}
  {0pt}
  {1mm}
  {-1em}
  {\bfseries}
}
\makeatother

\definecolor{winGreen}{HTML}{009900}
\definecolor{lossRed}{HTML}{CC0000}
\definecolor{winGreen2}{HTML}{1A8033}
\newcommand{\w}[1]{\textcolor{winGreen2}{#1}}

\providecommand{\authcount}[1]{}

\usepackage[pagebackref=true,breaklinks=true,colorlinks,bookmarks=false,allcolors=eccvblue]{hyperref}

\def\MethodName{StM}

\makeatletter
\renewcommand*{\@fnsymbol}[1]{\ensuremath{\ifcase#1\or \dagger \or \ddagger \or \mathsection \or \mathparagraph \or \|\else\@ctrerr\fi}}
\makeatother

\begin{document}

\title{Layer-Aware Video Composition via Split-then-Merge}

\author{
Ozgur Kara$^{1}$\thanks{Work done during an internship at Google.} \quad Yujia Chen$^{2}$ \quad Ming-Hsuan Yang$^{2}$ \\
James M. Rehg$^{1}$ \quad Wen-Sheng Chu$^{2}$\thanks{Joint last authors.} \quad Du Tran$^{2\ddagger}$
}
\authorrunning{O. Kara et al.}

\institute{%
$^{1}$University of Illinois Urbana-Champaign \quad $^{2}$Google
}

\maketitle

\vspace{-1.5em} 
\begin{center}
    \captionsetup{type=figure}
    \includegraphics[width=\textwidth]{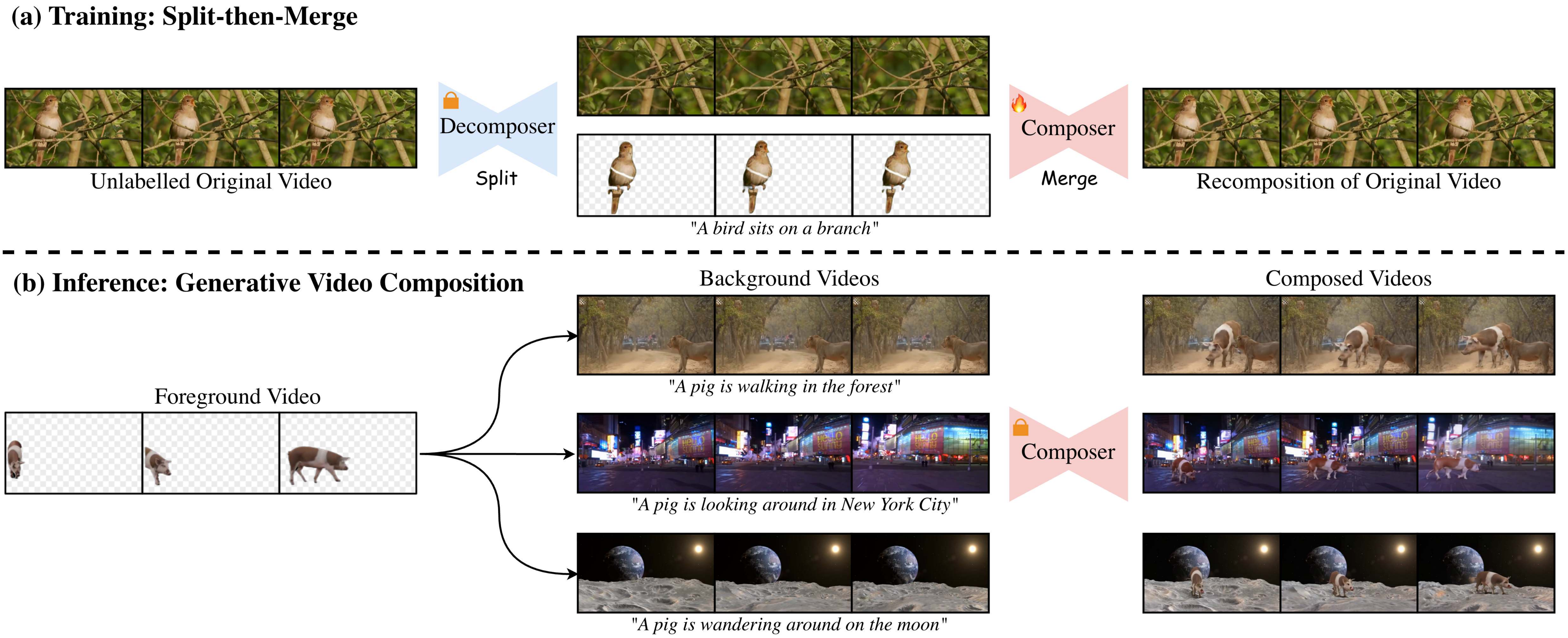}
    \vspace{-1.5ex}
    \caption{{\bf{{Video Composition via} Split-then-Merge.}}
    {\bf (a) Training:} The Decomposer \textit{splits} an unlabeled video into foreground and background layers and generates a caption, while the Composer learns to \textit{merge} them for reconstruction. 
    {\bf (b) Inference:} The Composer integrates a foreground video into novel background videos with affordance-aware placement (\eg, placing the pig on a rigid NYC walkway or uneven forest soil) with realistic harmonization of motion, lighting, and shadows.}
    \label{fig:teaser}
\end{center}
\vspace{-1.5em} 

\begin{abstract}
We present {\bf S}plit-{\bf t}hen-{\bf M}erge (\MethodName), a controllable generative video composition framework that minimizes reliance on annotated datasets and handcrafted rules.
Instead of requiring manual supervision, \MethodName~decomposes unlabeled videos into dynamic foreground and background layers. 
By self-composing these elements, the model learns to synthesize complex interactions between moving subjects and diverse scenes, capturing the intricate dynamics necessary for high-fidelity video generation. 
Specifically, \MethodName~introduces a transformation-aware training pipeline utilizing multi-layer fusion and augmentation to address affordance challenges in video composition. 
An identity-preservation loss further maintains foreground fidelity during blending. 
Extensive experiments show that \MethodName~outperforms state-of-the-art methods on both quantitative benchmarks and qualitative evaluations, including human studies and Vision-Language Model assessments. Finally, we release \MethodName-50K, the first multi-layer video dataset, to facilitate future research in generative video composition. 
Data, code, and more details are available at our \href{https://split-then-merge.github.io}{project page}.
\end{abstract}

\vspace{-5ex}
\section{Introduction}
\label{sec:intro}
\vspace{-1ex}

\begin{figure*}[t]
    \centering
    \includegraphics[width=\linewidth]{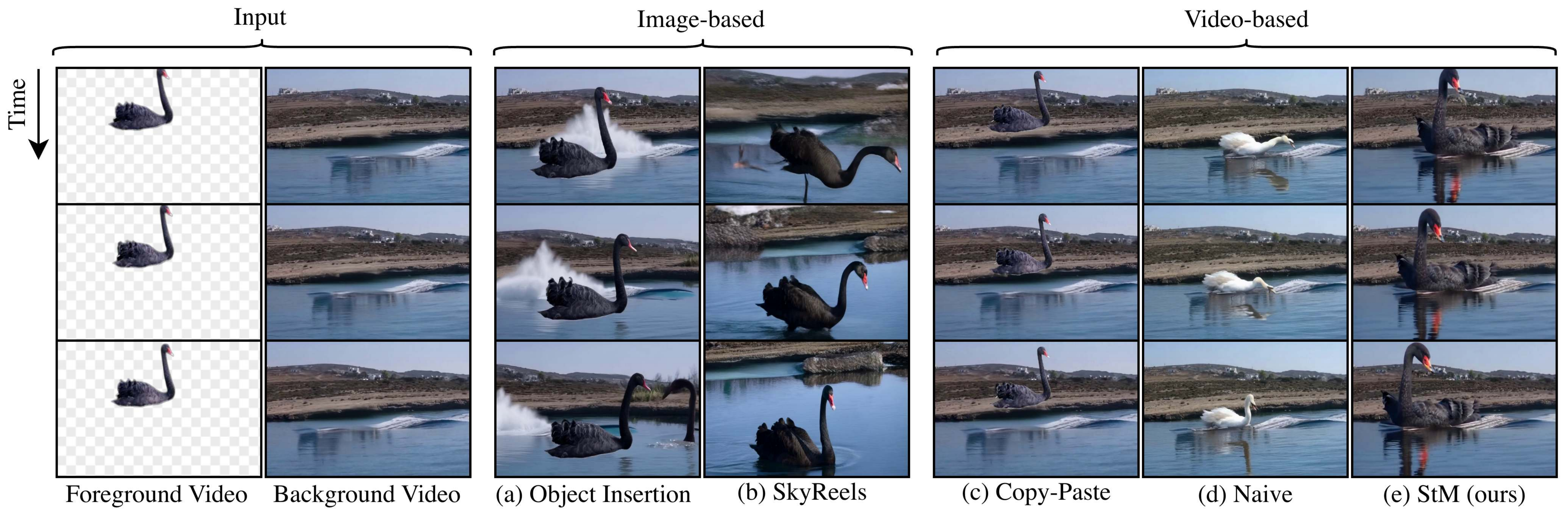}
    \vspace{-5ex}
    \caption{\textbf{{Video Composition.}}
    Given input foreground and background videos, image-based methods (a)--(b) use only the first frame, while (c)--(e) take full video inputs. 
    (a) Object insertion \cite{wu2025qwenimagetechnicalreport} followed by Image-to-Video (I2V) and (b) end-to-end I2V composition SkyReels~\cite{fei2025skyreels} fails to retain motion due to lack of video access.
    (c) Manual copy-paste preserves motion but violates affordance (swan placed on ground). 
    (d) Naive generative composition yields appearance and motion drift (\eg, black swan turns white). 
    (e) Our method preserves identity and motion, and achieves affordance-aware placement with realistic blending (swan placed in water with wave and shadows).
    }
    \label{fig:motivation}
\end{figure*}

Recent advances in diffusion models~\cite{ho2020denoising,song2021denoising,rombach2022high,peebles2023scalable,zhang2023adding} have greatly improved video realism. 
However, controllability remains largely limited to text~\cite{kara2024rave,geyer2024tokenflow,yang2025cogvideox,hong2023cogvideo,bar2024lumiere,gupta2024photorealistic} and image conditioning~\cite{guo2024i2v,ren2024consistiv,shi2024motion,namekata2025sgiv,ouyang2024i2vedit,feng2025ivcontrolcamera,zhao2024magdiff,jiang2024videobooth}, which lack precision for precise composition. 
While structured signals like pose~\cite{ma2024follow,chang2024magicpose,karras2023dreampose,hu2024animate} or motion~\cite{mou2024revideo,geng2025motion,burgert2025go} offer stronger control, most methods focus on animating static images rather than flexible synthesis. 
Professional content creation relies on layer-based compositing~\cite{brinkmann2008art,wright2013digital}, motivating \textit{generative video composition}: synthesizing a coherent video by integrating a dynamic foreground with a separate background using generative models, as depicted in Fig.~\ref{fig:teaser}(a). 
Unlike classical composition~\cite{Chen2013Motion}, which emphasizes matting and blending, the generative setting allows adaptive adjustments to foreground to harmonize with new lighting and camera dynamics. 
As shown in Fig.~\ref{fig:teaser}(b), the foreground animal's motion, appearance, and shadows are adapted to match diverse backgrounds.

One natural question is: ``\textit{How challenging is this task, and can it be addressed by existing methods?}'' 
A straightforward baseline combines image-based object insertion with an image-to-video (I2V) model. 
Specifically, the first frames of the foreground and background videos are composed using a state-of-the-art insertion method~\cite{wu2025qwenimagetechnicalreport}, and the resulting image is provided to an I2V model~\cite{yang2025cogvideox} for video generation. 
As shown in Fig.~\ref{fig:motivation}(a), this pipeline discards the temporal information of the foreground video, requiring the I2V model to infer motion solely from text. 
More recent approaches, such as SkyReels~\cite{fei2025skyreels}, which directly compose videos from foreground and background images, exhibit a similar limitation due to the absence of explicit motion cues (Fig.~\ref{fig:motivation}(b)).

An ensuing question is ``\textit{Can this task be addressed by extending methods to video inputs?}'' 
The answer is negative. 
A naive copy-and-paste strategy lacks affordance awareness and often yields implausible placements. 
As shown in Fig.~\ref{fig:motivation}(c), the swan is incorrectly positioned on the ground rather than in water. 
Fine-tuning a state-of-the-art video generator~\cite{yang2025cogvideox} with sufficient data is another attempt. 
However, naively trained models remain prone to affordance violations due to shortcut learning and fail to preserve motion and appearance consistency. 
Fig.~\ref{fig:motivation}(d) shows the swan's appearance shifting to white, violating foreground identity preservation. 
These results indicate that generative video composition is non-trivial and requires dedicated designs beyond off-the-shelf adaptations.

In this paper, we propose \textbf{S}plit-\textbf{t}hen-\textbf{M}erge (\MethodName), a data-driven framework for video composition requiring \emph{no} manual annotation (Fig.~\ref{fig:teaser}). 
Our key idea is to decompose unlabeled videos into foreground-background layers and train the model to reconstruct them, enabling scalable self-composition on large datasets (\eg, Panda-70M~\cite{chen2024panda}). 
Unlike application-specific editing methods (\eg, personalization~\cite{chen2025videoalchemist}, object swapping~\cite{gu2024videoswap}, motion control~\cite{tu2025videoanydoor,mou2024revideo,geng2025motion,burgert2025go}), \MethodName~is fully data-driven and generalizes across objects given a reliable Decomposer. 
We introduce two key components: (1) \emph{a transformation-aware training pipeline} that prevents shortcut learning and promotes affordance awareness, and (2) \emph{an identity-preservation loss} that maintains foreground identity while ensuring seamless background integration. 
In addition, we propose quantitative metrics for appearance and motion consistency, and leverage Vision-Language Models (VLMs) as automated judges to complement human evaluation. 
Our key contributions are:
\begin{compactenum} 
    \item We propose \MethodName, a scalable framework for generative video composition that eliminates manual annotations. \MethodName\ includes two novel components: the transformation-aware training pipeline and the identity-preservation loss.
    \item We release \MethodName-50K, the first multi-layer video dataset created through an automatic pipeline, which enables training future layer-based video generation methods.
    \item We present a comprehensive evaluation including strong baselines, new quantitative metrics, user studies, and VLM-based judges, demonstrate that \MethodName\ substantially outperforms existing methods. 
\end{compactenum}

\vspace{-1ex}
\section{Related Work}
\label{sec:rel_works}
\vspace{-1ex}

\paragraph{Controllable Video Generation.}
The success of diffusion models in image synthesis~\cite{ho2020denoising,song2021denoising,rombach2022high} has driven rapid progress in video generation, shifting emphasis from visual quality to controllability. 
Text-guided models~\cite{kara2024rave,geyer2024tokenflow,yang2025cogvideox,hong2023cogvideo,bar2024lumiere,gupta2024photorealistic,khachatryan2023text2video} provide high-level semantic guidance but lack precise control over identity and complex motion. 
Image-to-video (I2V) models~\cite{guo2024i2v,ren2024consistiv,shi2024motion,namekata2025sgiv,ouyang2024i2vedit,feng2025ivcontrolcamera,zhao2024magdiff,jiang2024videobooth} improve identity preservation via reference images, yet their static conditioning fails to capture temporal dynamics. 
Structural controls, such as pose~\cite{ma2024follow,chang2024magicpose,karras2023dreampose,hu2024animate} or spatial signals~\cite{guo2024sparsectrl,pmlr-v235-wang24cr,namekata2025sgiv,wu2024motionbooth,wu2024draganything,zhang2024controlvideo}, offer fine-grained guidance but remain limited to image-level animation. 
Consequently, existing methods are inherently image-centric, controlling isolated attributes rather than full video dynamics. 
In contrast, we tackle affordance-aware composition that integrates identity and motion from a foreground video with the dynamics of a background video.

\paragraph{Image and Video Composition.} 
Existing composition methods mainly rely on pixel-level blending techniques, such as color transfer~\cite{reinhard2001color,pitie2005n}, harmonization~\cite{cohenor2006color}, and gradient-domain pasting~\cite{jia2006drag}, to enhance realism~\cite{lalonde2007using,xue2012understanding}. 
Video extensions incorporate motion cues~\cite{Chen2013Motion,Wang2019Illumination} but lack generative capabilities. 
Generative harmonization of two dynamic video layers remains largely unexplored. 
Common baselines first insert objects at the image level (\eg, PbE~\cite{yang2023paint}, AnyDoor~\cite{chen2024anydoor}, Qwen-Edit~\cite{wu2025qwenimagetechnicalreport}), then animate the result with an I2V model; SkyReels~\cite{fei2025skyreels} follows a similar paradigm. 
However, these methods discard the original foreground motion by conditioning on static images. 
Other approaches differ in scope: AnyV2V~\cite{ku2024anyvv} propagates single-image edits, and VideoAnyDoor~\cite{tu2025videoanydoor} inserts a static image into a video. 
The most related work, LayerFlow~\cite{ji2025layerflow}, generates all layers from text. 
In contrast, \MethodName\ composes two existing video layers, explicitly preserving both their identity and motion.

\vspace{-1ex}
\section{Video Composition via Split-then-Merge}
\label{sec:methodology}
\vspace{-1ex}

\subsection{Generative Video Composition}
\vspace{-1ex}

Generative video composition aims to synthesize a coherent video by integrating a subject---together with its appearance and motion---from one source video into the dynamic scene of another. 
Formally, the inputs include: (i) a foreground video $V_{fg} \in [0,255]^{T \times H \times W}$ (omitting the color channel for simplicity), containing a primary subject and a binary mask $M_{fg} \in \{0,1\}^{T \times H \times W}$ associated with it; (ii) a background video $V_{bg}$ providing scene context; and (iii) a text prompt $\mathcal{T}$ describing the target scene. 
The objective is to generate $V_{pred}$ that seamlessly integrates the subject from $V_{fg}$ into $V_{bg}$ while preserving visual and motion consistency.

This task poses two challenges. 
First, \emph{layer consistency} requires preserving foreground appearance and motion (\eg, subject dynamics) as well as background scene and camera motion to avoid pasted-on artifacts. 
Second, \emph{affordance awareness} requires semantic understanding to ensure physically plausible placement and interaction (\eg, placing a car on a road rather than in the sky).

\subsection{Split-then-Merge Framework}
Split-then-Merge (\MethodName) is a generic data-driven framework for generative video composition based on \textit{self-composition}. 
The key idea is to use off-the-shelf models to \textit{split} an unlabeled video $V_{org}$ into layers and train a model to \textit{merge} them for reconstruction (Fig.~\ref{fig:teaser}(a)). 
Without human annotation, the \MethodName~Decomposer automatically separates any unlabeled video into a foreground subject and background scene, and generates a text caption. 
This process converts raw videos into training samples, enabling large-scale multi-layer dataset construction. 
A diffusion-based generative model (\ie, the Composer) is then trained to \textit{merge} the layers back into the original coherent video.

This approach differs from conventional generative video and computational photography methods, which rely on specialized algorithms or domain heuristics~\cite{pan2024actanywhere,ji2025layerflow,ku2024anyvv,gu2024videoswap,geyer2024tokenflow,baliah2025realistic}. 
Such methods are often confined to narrow domains (\eg, face swapping~\cite{baliah2025realistic,luo2025canonswap,shao2025vividface}), limiting scalability to diverse in-the-wild content. 
By casting the task as inverting a decomposition process, our framework avoids these constraints and learns realistic video composition directly from data.

\subsection{Decomposer: Splitting Videos into Layers}
\label{sec:decomposer}

\begin{figure}[t]
\centering
\includegraphics[width=\columnwidth]{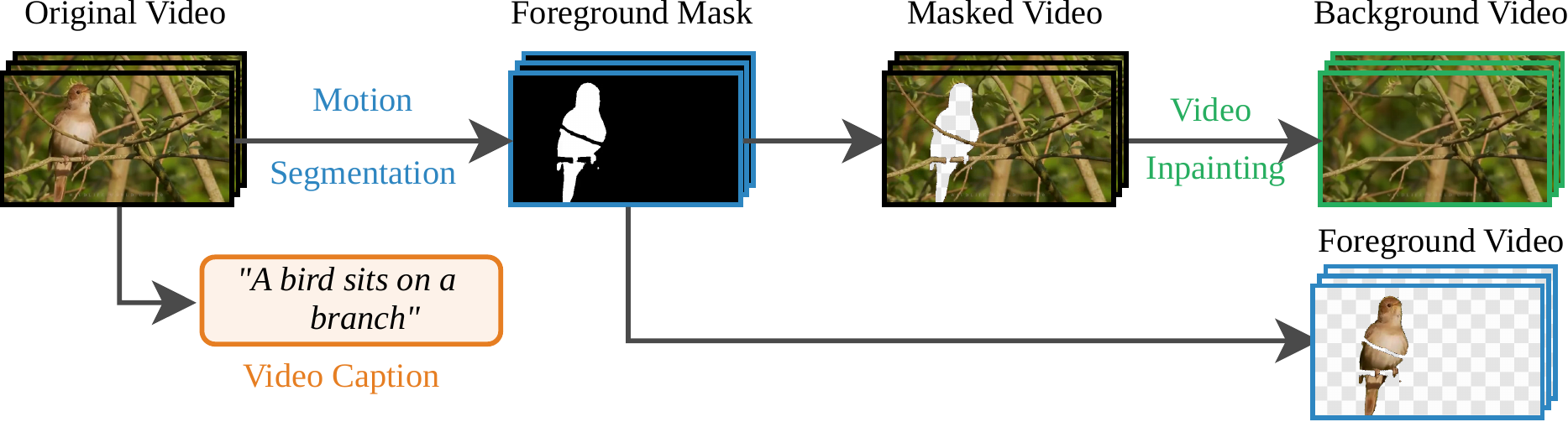}
    \caption{{\textbf{StM Decomposer.}} The StM Decomposer integrates off-the-shelf models to split unlabeled videos. First, motion segmentation generates a foreground mask, which is used to extract the foreground layer. An inpainting model then fills the ``holes" in the masked background video. Finally, a video captioning model generates a descriptive text caption for the original video.}
    \label{fig:decomposer}
\end{figure}
A key challenge in generative video composition is the lack of large-scale multi-layer video datasets. 
To address this, we design an automated pipeline that decomposes videos into four components: a caption, a foreground layer, its mask, and a background layer. 
As shown in Fig.~\ref{fig:decomposer}, the \MethodName~Decomposer first employs a pre-trained video-language model~\cite{wang2024internvid} to generate captions. 
An automatic motion segmentation model (\ie, Segment-Any-Motion~\cite{huang2025segment}) then extracts the primary moving subject and its mask. 
The masked region is removed, and a state-of-the-art video inpainting model~\cite{zi2025minimax} reconstructs a coherent background. 

\begin{figure*}[t]
\centering
\includegraphics[width=\textwidth]{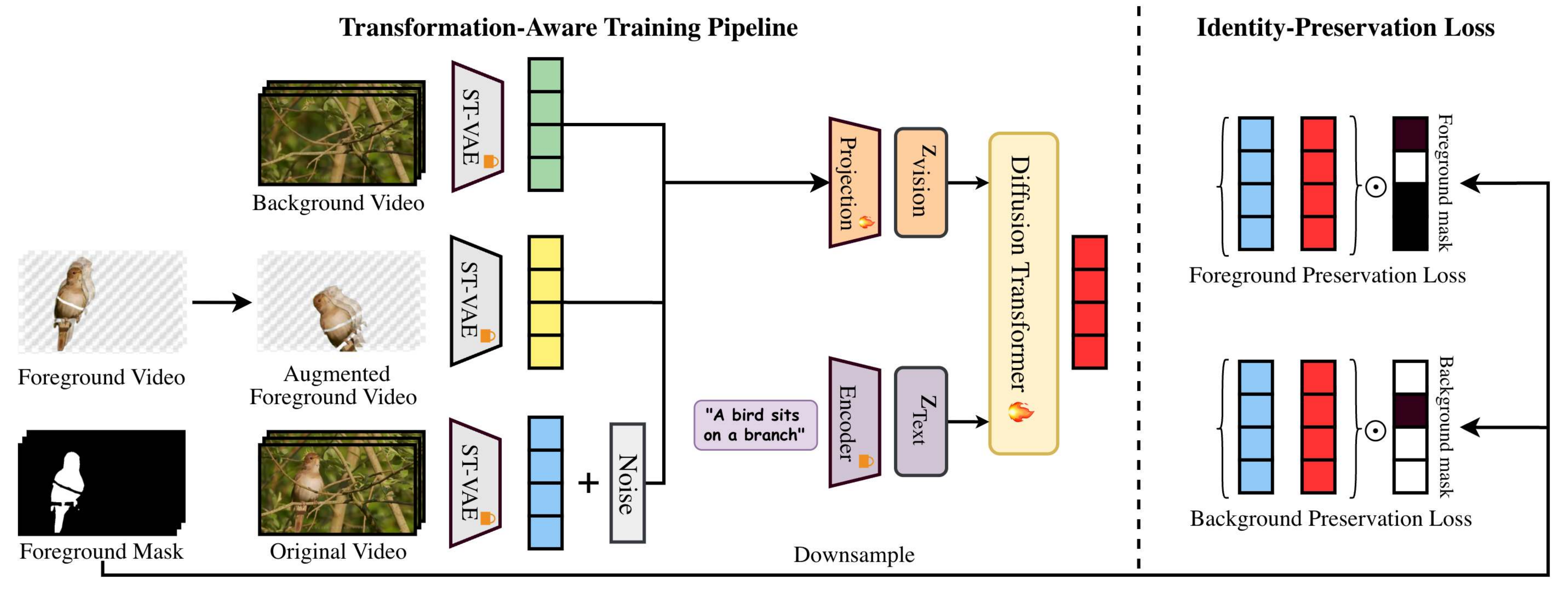}
    \caption{{\textbf{StM Composer Training}}. 
    The Composer is trained to reconstruct a ground-truth video latent from foreground, background, and text inputs. 
    First, the foreground video is augmented, and all video inputs (augmented foreground, background, ground truth) are encoded into latents by a frozen Space-Time (ST) VAE. 
    The text prompt is encoded as $Z_{\text{text}}$. A noisy ground-truth latent (blue) is fused with background (green) and augmented foreground (yellow) latents via a projection layer to produce the visual representation $Z_{\text{vision}}$. 
    A Diffusion Transformer then processes $Z_{\text{vision}}$ and $Z_{\text{text}}$ to predict a composed latent (red). 
    The identity-preservation loss comprises two weighted sub-losses comparing the prediction (red) against the ground truth (blue) using foreground- and background-aware masking. 
    }
    \label{fig:recomposer}
\end{figure*}

\subsection{Composer: Learning to Merge}
\label{sec:composer}
The \MethodName~Composer builds upon a latent diffusion transformer~\cite{yang2025cogvideox} for text-to-video generation. 
It consists of a pre-trained space-time VAE encoder-decoder ($E_v, D_v$), a text encoder ($E_T$), and a Diffusion Transformer (DiT) backbone $\mathbf{f}(x_t;\theta)$, where $x_t$ denotes the latent input, $\theta$ represents the model parameters, and $t$ is the diffusion timestep. 
The standard objective is to recover the clean latent from a noise-added latent conditioned on text embeddings. 
We adapt this architecture for video composition via a transformation-aware training pipeline with three components: multi-layer conditional fusion, transformation-aware augmentation, and an identity-preservation loss.

\paragraph{Multi-layer Conditional Fusion.} 
Fig.~\ref{fig:recomposer} illustrates the \MethodName~Composer architecture. 
The foreground video $V_{\text{fg}}$ is first augmented to obtain $\tilde{V}_{\text{fg}}$. 
The original video, background video, and augmented foreground are encoded by $E_v$ to produce latent representations $\mathbf{z}_{\text{org}}, \mathbf{z}_{\text{bg}}, \tilde{\mathbf{z}}_{\text{fg}}$, where $\mathbf{z}_i = E_v(V_i)$. 
During diffusion, noise $\epsilon_t$ is added to $\mathbf{z}_{\text{org}}$ following diffusion schedule $\bar{\alpha}_t$, and all visual latents are fused via channel-wise concatenation followed by an MLP:
\begin{equation}
    \mathbf{z}_{\text{vision},t} = \text{MLP}(\text{Concat}_{\text{channel}}(\sqrt{\bar{\alpha}_t} \mathbf{z}_{\text{org}} + \sqrt{1 - \bar{\alpha}_t} \epsilon, \tilde{\mathbf{z}}_{\text{fg}}, \mathbf{z}_{\text{bg}})).
\end{equation}
This \emph{channel-level conditioning} provides dense spatio-temporally aligned guidance, in contrast to token-based or cross-attention conditioning. 
By integrating foreground, background, and noisy input at each latent location, it enables joint reasoning over all visual components and facilitates precise recomposition.
The text prompt $\mathcal{T}$ is encoded as $\mathbf{z}_{\text{text}} = E_T(\mathcal{T})$. 
The final transformer input is $\mathbf{x}_t = \text{Concat}_{\text{token}}(\text{Tokens}(\mathbf{z}_{\text{vision},t}), \mathbf{z}_{\text{text}})$,
which is fed into the DiT model to predict $\mathbf{z}_t = \mathbf{f}(\mathbf{x}_t;\theta)$, reconstructing the clean latent $\mathbf{z}_0 \overset{\Delta}{=} \mathbf{z}_{\text{org}}$.

\paragraph{Transformation-Aware~Augmentation.}
To prevent ``copy-paste'' shortcuts and promote genuine compositional reasoning, we introduce a targeted augmentation strategy. 
The key idea is to increase training difficulty by applying random transformations solely to the conditioning foreground $V_{fg}$. 
The model must therefore \textit{\textbf{invert}} these transformations to correctly and plausibly place the subject within the background. 
Specifically, we apply spatial augmentations (random horizontal flipping, cropping, and resizing) and photometric changes (color jittering) to the entire video. 
By perturbing the subject's orientation, scale, position, and color, we prevent memorization of fixed layouts and encourage true \textit{\textbf{affordance awareness}} and \textit{\textbf{color harmony}}. 
This compels the model to learn placement and harmonization principles rather than replicate patterns.

\paragraph{Identity-Preservation Loss.} 
The standard latent diffusion objective is:
\begin{equation}
    \mathcal{L}_{\text{recon}} = \mathbb{E}_{\mathbf{z}_0, t} \left[\lVert \mathbf{z}_0 - \mathbf{f}(\mathbf{x}_t, t; \theta)\rVert^2 \right],
\end{equation}
where we omit the constant weight $w_t$ for simplicity. 
This loss treats all latent locations uniformly, as $E_v$ preserves spatio-temporal locality and the $\ell_2$ objective assigns equal weight across regions. 
However, video composition must balance foreground identity with visual harmony. 
Over-optimizing for harmony can distort identity (as shown in Fig.~\ref{fig:motivation}(d)), whereas strictly enforcing it often yields disjoint compositions.

We introduce an identity-preservation loss to balance foreground identity retention and overall harmony. 
Given the training-time foreground mask $\mathbf{M}$, we partition the latent space into foreground and background regions with separate weights, yielding:
\begin{align}
\mathcal{L}_{\text{fg}} &= \mathbb{E}_{\mathbf{z}_0, t} \left[ 
\frac{\sum \big( (\mathbf{z}_0 - \mathbf{f}(\mathbf{x}_t, t; \theta))^2 \odot \mathbf{M} \big)}{\sum \mathbf{M}} \right], \nonumber \\
\mathcal{L}_{\text{bg}} &= \mathbb{E}_{\mathbf{z}_0, t} \left[ 
\frac{\sum \big( (\mathbf{z}_0 - \mathbf{f}(\mathbf{x}_t, t; \theta))^2 \odot (1-\mathbf{M}) \big)}{\sum (1-\mathbf{M})} \right].
\end{align}
Both terms are normalized by their respective areas to reduce sensitivity to object size. 
The final loss is a weighted combination:
\begin{align}
\mathcal{L}_{\text{final}} = \alpha \mathcal{L}_{\text{fg}} + (1-\alpha)\mathcal{L}_{\text{bg}}.
\end{align}
This formulation explicitly controls the trade-off between identity preservation and compositional harmony.

\paragraph{Inference.} Inference follows the training procedure with three differences: 
(1) the ground-truth latent is replaced by a sampled noise latent $\epsilon$,
(2) foreground augmentation is disabled,
and (3) the space-time decoder ($D_v$) maps the predicted latent back to pixel space.

\vspace{-1ex}
\section{Experiments}
\label{sec:experimentation}
\vspace{-1ex}

\subsection{Implementation Details}
\vspace{-1ex}

\paragraph{Decomposer.}
Our Decomposer (Section~\ref{sec:decomposer}) processes unlabeled videos with off-the-shelf models: InternVL~\cite{chen2024internvl} for captioning, Segment-Any-Motion~\cite{huang2025segment} for foreground masks, and MiniMax Remover~\cite{zi2025minimax} for background inpainting.

\paragraph{Constructing \MethodName-50K dataset.}
Because there are no existing large-scale video datasets with comprehensive multi-layer annotations, creating this resource required rigorous curation and strategic sampling across diverse sources to balance semantic categories and capture a wide spectrum of complex motion dynamics. For training, we purposefully sourced videos from multiple datasets to encompass a diverse set of subjects: animals (Animal Kingdom~\cite{ng2022animal}), humans (Panda-70M~\cite{chen2024panda}), and general objects (Youtube-VOS~\cite{xu2018youtube}, LVOS~\cite{hong2023lvos}). We then applied our automatic pipeline to decompose these sampled videos into layers, resulting in \MethodName-50K, a training dataset of 50,000 processed clips.

For evaluation, we rely exclusively on the DAVIS dataset~\cite{perazzi2016benchmark}. We curated 93 unseen triplets (foreground, background, text) where the foreground and background originate from different videos. These triplets span diverse semantic categories---including animals, human actions, and rigid objects---to rigorously test compositional generalization and ensure a comprehensive evaluation across varying motion complexities. Note this evaluation set size is comparable to standard video generation benchmarks~\cite{ji2025layerflow, ku2024anyvv, fei2025skyreels} ranging from 50--100 videos.

\begin{table*}[t]
\caption{\textbf{\textit{Quantitative Evaluation Metrics.}} Each metric computes the similarity or distance $f(.,.)$ between the representation of the ground truth ($\mathbf{r}_{\text{gt}}$) and the corresponding predicted ($\mathbf{r}_{\text{pred}}$) layers. Here $\mathbf{v}_{\text{gt}}$ and $\mathbf{v}_{\text{pred}}$ denote the ground truth and generated output videos, respectively, with superscripts $\text{FG}$ and $\text{BG}$ indicating the segmented foreground and background layers, and $\mathbf{t}$ denotes the guidance text prompt. ViCLIP~\cite{wang2024internvid}, VideoSwin~\cite{liu2022video}, and OptFlow~\cite{jaegle2022perceiver} denote methods mapping video layers (or text) into representations.}
    \label{tab:eval_metrics}
    \vspace{-3mm}
    \centering
    \scriptsize
    \renewcommand{\arraystretch}{1.15}
    \begin{tabularx}{\textwidth}{
      @{}
      >{\raggedright\arraybackslash}p{5.2cm}
      >{\centering\arraybackslash}p{1.8cm}
      >{\centering\arraybackslash}X
      >{\centering\arraybackslash}X
      @{}
    }
    \toprule
    \textbf{Metric} & $f(\cdot,\cdot)$ & $\mathbf{r}_{\text{gt}}$ & $\mathbf{r}_{\text{pred}}$ \\
    \midrule
    \rowcolor{RoyalBlue!6}
    \textbf{(M1)} FG ID Preservation {\scriptsize Consistency of the FG subject.}
    & CosSim $\uparrow$
    & ViCLIP($\mathbf{v}_{\text{gt}}^{\text{FG}}$)
    & ViCLIP($\mathbf{v}_{\text{pred}}^{\text{FG}}$) \\
    \textbf{(M2)} BG ID Preservation {\scriptsize Consistency of the BG scene.}
    & CosSim $\uparrow$
    & ViCLIP($\mathbf{v}_{\text{gt}}^{\text{BG}}$)
    & ViCLIP($\mathbf{v}_{\text{pred}}^{\text{BG}}$) \\
    \rowcolor{RoyalBlue!6}
    \textbf{(M3)} Semantic Action Alignment {\scriptsize Fidelity of FG action to the ground-truth.}
    & KL Div.\ $\downarrow$
    & VideoSwin($\mathbf{v}_{\text{gt}}^{\text{FG}}$)
    & VideoSwin($\mathbf{v}_{\text{pred}}^{\text{FG}}$) \\
    \textbf{(M4)} BG Motion Alignment {\scriptsize Consistency of BG motion dynamics.}
    & MSE $\downarrow$
    & OptFlow($\mathbf{v}_{\text{gt}}^{\text{BG}}$)
    & OptFlow($\mathbf{v}_{\text{pred}}^{\text{BG}}$) \\
    \rowcolor{RoyalBlue!6}
    \textbf{(M5)} Textual Alignment \newline {\scriptsize Alignment between video and text.}
    & CosSim $\uparrow$
    & ViCLIP($\mathbf{t}$)
    & ViCLIP($\mathbf{v}_{\text{pred}}$) \\
    \bottomrule
    \end{tabularx}
\end{table*}

\begin{figure*}[t!]
    \centering
    \includegraphics[width=\textwidth]{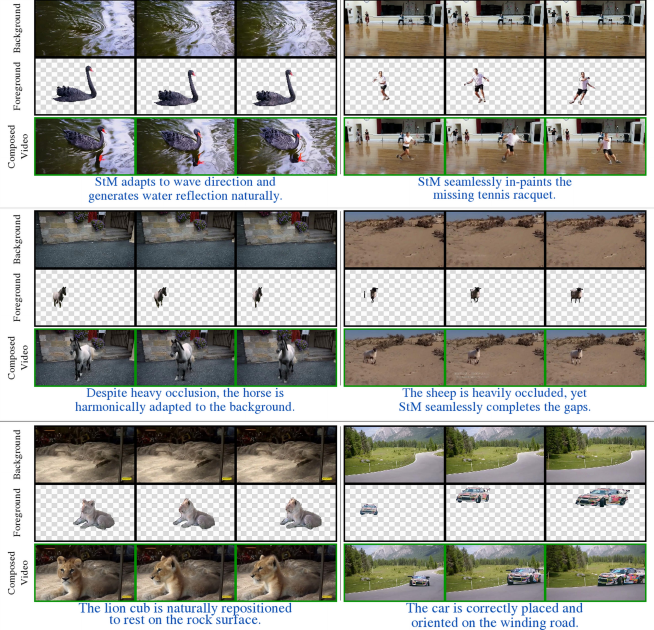}
    \caption{{\textbf{Qualitative Results.}} \MethodName\ demonstrates robust motion preservation and affordance awareness. Characteristic actions are faithfully retained (\eg, cheetah, tennis player). Subjects are plausibly integrated rather than ``pasted,'' as seen with the swan and duck adapted to water flow, and the lion and race car harmonized with appropriate scene lighting and placement. Full videos are available on our \href{https://split-then-merge.github.io}{project page}.}
    \label{fig:qual-res}
\end{figure*}

\paragraph{Composer.}
We adopt CogVideoX-I2V~\cite{yang2025cogvideox} as the base model, initialize with pre-trained weights, and fine-tune for 20K iterations. 
Training uses 16 NVIDIA H100 GPUs (batch size $64$) with \texttt{bf16} precision. 
We employ AdamW~\cite{loshchilov2018decoupled} ($\beta_1\!=\!0.9$, $\beta_2\!=\!0.95$, $\epsilon\!=\!1\mathrm{e}{-8}$, weight decay $1\mathrm{e}{-4}$, max grad norm $1.0$) and a cosine scheduler (base LR $5\mathrm{e}{-6}$, 1K warm-up). 
Videos are resized to $49 \!\times\! 480 \!\times\! 720$, and the identity-preservation weight is $\alpha=0.5$.

To mitigate shortcut learning, we apply the following temporally consistent augmentations sequentially to the foreground video: 
(1) random horizontal flipping ($p\!=\!0.7$); 
(2) random resized cropping ($p\!=\!0.7$, scale [0.5, 2.0], preserving 90\% foreground); 
(3) color jittering ($p\!=\!0.2$) exclusively on the masked foreground to perturb brightness, contrast, saturation, and hue within [0, 0.2].

\subsection{Evaluation Setups}

\begin{table*}[t!]
\centering
\caption{\textbf{{Quantitative Evaluation and Ablation Study.}} Comparison with baseline methods (top) and ablations of each component (bottom) by removing them from our full model. We report \textbf{M1} (FG ID Preservation), \textbf{M2} (BG ID Preservation), \textbf{M3} (Semantic Action Alignment), \textbf{M4} (BG Motion Alignment), and \textbf{M5} (Textual Alignment). Metric directions are marked ($\uparrow, \downarrow$). M1, M2, and M5 scores are multiplied by 100. \textbf{Bold} indicates the best performance, and \underline{underline} is the second best. Percentage changes in the ablation study are relative to our full model.}
\label{tab:quant_eval_and_ablation}
\vspace{-1ex}
\resizebox{1.0\columnwidth}{!}{%
\setlength{\tabcolsep}{6pt}
\renewcommand{\arraystretch}{1.1}
\begin{tabular}{@{}cc ccccc@{}}
\toprule
\multicolumn{2}{@{}l}{\multirow{2}{*}{\textbf{Method}}} & \multicolumn{2}{c}{\textbf{ID Preservation}} & \multicolumn{2}{c}{\textbf{Action / Motion Align.}} & \multirow{2}{*}{\shortstack{\textbf{Text Align.}\\(M5) $\uparrow$}} \\
\cmidrule(lr){3-4} \cmidrule(lr){5-6}
\multicolumn{2}{@{}l}{} & FG (M1) $\uparrow$ & BG (M2) $\uparrow$ & Sem.\ Action (M3) $\downarrow$ & BG Motion (M4) $\downarrow$ & \\
\midrule
\rowcolor{gray!6}
\multicolumn{2}{@{}l}{Copy-Paste + I2V} & \underline{83.08} & \underline{85.02} & \underline{1.61} & 184.59 & 19.21 \\
\multicolumn{2}{@{}l}{PBE + I2V}        & 73.52 & 80.19 & 2.47 & 98.08 & 19.41 \\
\rowcolor{gray!6}
\multicolumn{2}{@{}l}{Qwen + I2V}       & 82.02 & 72.38 & 1.71 & 74.77 & \underline{24.20} \\
\multicolumn{2}{@{}l}{SkyReels}         & 80.24 & 75.24 & 1.75 & 279.23 & \textbf{24.40} \\
\rowcolor{gray!6}
\multicolumn{2}{@{}l}{AnyV2V}           & 77.73 & 56.36 & 1.70 & 154.97 & 24.13 \\
\multicolumn{2}{@{}l}{LayerFlow-FG}     & 83.02 & 60.21 & 1.47 & 143.47 & 18.74 \\
\rowcolor{gray!6}
\multicolumn{2}{@{}l}{LayerFlow-BG}     & 50.27 & 91.73 & 2.35 & \underline{24.72} & 19.57 \\
\midrule
\rowcolor{blue!8}
\multicolumn{2}{@{}l}{\textbf{\MethodName} (ours)} & \textbf{84.82} & \textbf{92.88} & \textbf{1.22} & \textbf{16.36} & 19.81 \\
\midrule
\rowcolor{gray!12}
\textbf{Aug.} & \textbf{ID Loss} & \multicolumn{5}{c}{\textbf{Ablation Study}} \\
\midrule
\rowcolor{blue!8}
$\textcolor{winGreen}{\checkmark}$ & $\textcolor{winGreen}{\checkmark}$ & 84.82 & \textbf{92.88} & 1.22 & \textbf{16.36} & \textbf{19.81} \\
\rowcolor{gray!6}
$\textcolor{winGreen}{\checkmark}$ & $\textcolor{lossRed}{\times}$ 
  & 82.01 {\scriptsize \textcolor{lossRed}{(-3.3\%)}} 
  & 88.25 {\scriptsize \textcolor{lossRed}{(-5.0\%)}} 
  & 0.92 {\scriptsize \textcolor{winGreen}{(-24.6\%)}} 
  & 23.75 {\scriptsize \textcolor{lossRed}{(+45.2\%)}}
  & 16.42 {\scriptsize \textcolor{lossRed}{(-17.1\%)}} \\
$\textcolor{lossRed}{\times}$ & $\textcolor{winGreen}{\checkmark}$ 
  & 83.15 {\scriptsize \textcolor{lossRed}{(-2.0\%)}} 
  & 89.20 {\scriptsize \textcolor{lossRed}{(-4.0\%)}} 
  & 0.70 {\scriptsize \textcolor{winGreen}{(-42.6\%)}} 
  & 16.61 {\scriptsize \textcolor{lossRed}{(+1.5\%)}}
  & 16.56 {\scriptsize \textcolor{lossRed}{(-16.4\%)}} \\
$\textcolor{lossRed}{\times}$ & $\textcolor{lossRed}{\times}$ 
  & \textbf{90.39} {\scriptsize \textcolor{winGreen}{(+6.6\%)}} 
  & 84.76 {\scriptsize \textcolor{lossRed}{(-8.7\%)}} 
  & \textbf{0.77} {\scriptsize \textcolor{winGreen}{(-36.9\%)}} 
  & 18.59 {\scriptsize \textcolor{lossRed}{(+13.6\%)}}
  & 18.70 {\scriptsize \textcolor{lossRed}{(-5.6\%)}} \\
\bottomrule
\end{tabular}
} %
\vspace{2ex}
\end{table*}

\paragraph{Baselines.}
We evaluate \MethodName~against 7 SoTA methods in three groups: \textit{cascaded I2V}, \textit{video-centric}, and \textit{layer-aware} methods. 
To ensure a strictly fair comparison, cascaded I2V baselines utilize the exact same pre-trained CogVideoX-I2V backbone, inference steps, and classifier-free guidance scales as our StM Composer.
These include \textbf{Copy-Paste + I2V} (centered paste with half-scale bounding box), \textbf{PBE + I2V} using Paint-by-Example~\cite{yang2023paint}, and \textbf{Qwen + I2V} using Qwen-Edit~\cite{wu2025qwenimagetechnicalreport}. 
Video-centric baselines include \textbf{SkyReels}~\cite{fei2025skyreels}, which composes static images end-to-end, and \textbf{AnyV2V}~\cite{ku2024anyvv}, a tuning-free motion-aware editor. 
Layer-aware methods like \textbf{LayerFlow-FG} and \textbf{LayerFlow-BG}~\cite{ji2025layerflow} condition on foreground or background layers, respectively. 
Notably, cascaded I2V baselines and SkyReels cannot preserve original foreground motion due to reliance on static inputs.

\paragraph{Metrics.}
We evaluate all methods using five criteria through automated metrics and qualitative studies (user study and VLM-as-a-Judge). 
For quantitative evaluation, we use~\cite{huang2025segment} to decompose generated videos into foreground (FG) and background (BG) layers, which are compared with input FG/BG layers using the metrics in Table~\ref{tab:eval_metrics}. 
M1/M2 (FG/BG ID Preservation) measure appearance consistency; M3 (Semantic Action Alignment) evaluates FG semantic motion; M4 (BG Motion Alignment) measures BG pixel-level motion (camera and scene dynamics); M5 (Textual Alignment) measures fidelity of the composite video to the guidance text prompt. M1--M4 are purposely designed to evaluate video composition by directly measuring fidelity. 
M5 is a standard metric for image and video generation, which we provide for completeness.
For qualitative evaluation, we conduct a user study with 50 participants on Prolific~\cite{prolific} using 25 randomly sampled test cases, and employ Gemini 2.5 Pro~\cite{comanici2025gemini} as a VLM judge on the full test set (see Appendix~\ref{sec:vllm_judge}). 
The four criteria are: (i) \textit{\textbf{Identity Preservation (FG \& BG)}}; 
(ii) \textit{\textbf{Motion Alignment (FG \& BG)}}; 
(iii) \textit{\textbf{FG-BG Harmony}} (qualitative only); and 
(iv) \textit{\textbf{Overall Quality}} (qualitative only).

\subsection{Analysis}
\vspace{-1ex}

\begin{table*}[t]
\centering
\caption{\textbf{Pairwise Preference Study (Win Rates).} 
         We present a qualitative study where human evaluators and a VLM judge (Gemini 2.5 Pro) performed pairwise 
         comparisons between our method and a baseline. 
         The percentages reflect 
         our method's ``win rate'' (\ie, how often it was preferred over the baseline).
         Values in \textcolor{winGreen}{green} indicate our method is preferred (>50$\%$), while \textcolor{lossRed}{{\bf red}} indicates the baseline is preferred (<50$\%$).}
    \label{tab:quant_pref_merged}
    \resizebox{\textwidth}{!}{%
    \setlength{\tabcolsep}{4pt}
    \renewcommand{\arraystretch}{1.15}
\begin{tabular}{@{}l cccccc @{\hspace{10pt}} cccccc@{}}
\toprule
\multirow{3}{*}{\textbf{Ours vs.\ Baseline}} 
& \multicolumn{6}{c}{\textbf{User Study} (in \%)} 
& \multicolumn{6}{c}{\textbf{VLM Judge} (in \%)} \\
\cmidrule(lr){2-7} \cmidrule(lr){8-13} 
& \multicolumn{2}{c}{Identity Preserv.} & \multicolumn{2}{c}{Motion Align.} & \multirow{2}{*}{\shortstack{FG-BG \\ Harmony}} & \multirow{2}{*}{\shortstack{Overall \\ Quality}}
& \multicolumn{2}{c}{Identity Preserv.} & \multicolumn{2}{c}{Motion Align.} & \multirow{2}{*}{\shortstack{FG-BG \\ Harmony}} & \multirow{2}{*}{\shortstack{Overall \\ Quality}} \\
\cmidrule(lr){2-3} \cmidrule(lr){4-5} \cmidrule(lr){8-9} \cmidrule(lr){10-11} 
& FG & BG & FG & BG & &
& FG & BG & FG & BG & & \\
\midrule
\rowcolor{gray!6}
Copy-Paste + I2V & \textcolor{winGreen}{72.73} & \textcolor{winGreen}{83.04} & \textcolor{winGreen}{76.65} & \textcolor{winGreen}{84.90} & \textcolor{winGreen}{86.53} & \textcolor{winGreen}{86.50}
                 & \textcolor{winGreen}{57.30} & \textcolor{winGreen}{85.88} & \textcolor{winGreen}{65.17} & \textcolor{winGreen}{80.00} & \textcolor{winGreen}{91.95} & \textcolor{winGreen}{90.00} \\
PBE + I2V        & \textcolor{winGreen}{87.77} & \textcolor{winGreen}{88.00} & \textcolor{winGreen}{84.76} & \textcolor{winGreen}{88.06} & \textcolor{winGreen}{88.54} & \textcolor{winGreen}{88.50}
                 & \textcolor{winGreen}{87.78} & \textcolor{winGreen}{95.35} & \textcolor{winGreen}{84.27} & \textcolor{winGreen}{90.91} & \textcolor{winGreen}{90.80} & \textcolor{winGreen}{92.22} \\
\rowcolor{gray!6}
Qwen + I2V       & \textcolor{winGreen}{59.06} & \textcolor{winGreen}{73.56} & \textcolor{winGreen}{61.23} & \textcolor{winGreen}{70.44} & \textcolor{winGreen}{55.11} & \textcolor{winGreen}{55.10}
                 & \textcolor{winGreen}{52.17} & \textcolor{winGreen}{90.36} & \textcolor{winGreen}{52.17} & \textcolor{winGreen}{84.81} & \textcolor{winGreen}{54.95} & \textcolor{lossRed}{{\bf 46.73}} \\
SkyReels         & \textcolor{winGreen}{68.74} & \textcolor{winGreen}{82.38} & \textcolor{winGreen}{67.38} & \textcolor{winGreen}{79.59} & \textcolor{winGreen}{64.46} & \textcolor{winGreen}{64.50}
                 & \textcolor{winGreen}{57.14} & \textcolor{winGreen}{94.44} & \textcolor{winGreen}{61.11} & \textcolor{winGreen}{89.87} & \textcolor{winGreen}{69.66} & \textcolor{winGreen}{65.93} \\
\rowcolor{gray!6}
AnyV2V           & \textcolor{winGreen}{84.45} & \textcolor{winGreen}{87.46} & \textcolor{winGreen}{88.80} & \textcolor{winGreen}{88.72} & \textcolor{winGreen}{89.62} & \textcolor{winGreen}{89.60}
                 & \textcolor{winGreen}{96.10} & \textcolor{winGreen}{100.00} & \textcolor{winGreen}{98.72} & \textcolor{winGreen}{98.75} & \textcolor{winGreen}{95.00} & \textcolor{winGreen}{100.00} \\
LayerFlow-FG     & \textcolor{winGreen}{51.00} & \textcolor{winGreen}{94.79} & \textcolor{winGreen}{54.15} & \textcolor{winGreen}{91.55} & \textcolor{winGreen}{86.76} & \textcolor{winGreen}{88.71} 
                 & \textcolor{winGreen}{52.04} & \textcolor{winGreen}{100.00} & \textcolor{winGreen}{62.24} & \textcolor{winGreen}{95.92} & \textcolor{winGreen}{87.76} & \textcolor{winGreen}{90.82} \\                                 
\rowcolor{gray!6}
LayerFlow-BG     & \textcolor{winGreen}{94.89} & \textcolor{winGreen}{83.08} & \textcolor{winGreen}{94.32} & \textcolor{winGreen}{84.12} & \textcolor{winGreen}{93.36} & \textcolor{winGreen}{95.89}  
                 & \textcolor{winGreen}{93.88} & \textcolor{winGreen}{56.12} & \textcolor{winGreen}{96.94} & \textcolor{winGreen}{77.55} & \textcolor{winGreen}{97.96} & \textcolor{winGreen}{95.92} \\
\bottomrule
\end{tabular}%
} 
\vspace{1ex}
\end{table*}

\paragraph{Quantitative Evaluation.}
Table~\ref{tab:quant_eval_and_ablation} shows the quantitative advantages of \MethodName. 
For \textit{\textbf{Identity Preservation}}, \MethodName\ achieves the highest scores for both foreground (M1: 84.82) and background (M2: 92.88), indicating reduced appearance drift compared to I2V baselines that infer dynamics from a single frame (\eg, PBE + I2V). 
The gap widens in \textit{\textbf{Action \& Motion Alignment}} (lower is better). 
\MethodName\ attains the lowest FG action error (M3: 1.22), outperforming baselines (1.47--2.47) that hallucinate motion. 
For BG motion alignment, our method scores M4: 16.36, substantially better than competitors (24.72--279.23), which often fail to preserve camera and scene dynamics and generate static backgrounds. 

For \textit{\textbf{Textual Alignment}} (M5), our method scores 19.81, performing similarly to layer-aware baselines LayerFlow (18.74--19.57). 
This is expected, as \MethodName~relies primarily on foreground-background layers for control, with the text prompt providing complementary guidance to harmonize the composite rather than driving generation from scratch. 
Methods that achieve higher M5 scores (\eg, SkyReels, Qwen + I2V) operate in a fully text-driven regime where the prompt is the sole generative signal, naturally yielding stronger text alignment at the cost of identity and motion fidelity. 
In addition, because we introduce {\em no new parameters beyond a projection layer}, our inference time remains identical to the CogVideoX-I2V baseline.
These quantitative results align with qualitative failures observed in baselines, such as a lack of affordance awareness in Copy-Paste + I2V and content inconsistency in Qwen + I2V.

\paragraph{User Study / VLM as a Judge.}
We conduct qualitative preference studies (Table~\ref{tab:quant_pref_merged}) to complement automated evaluations (Table~\ref{tab:quant_eval_and_ablation}) and assess attributes beyond quantitative metrics, including FG-BG Harmony and Overall Quality. 
For Identity Preservation, both human and VLM judgments align with automated results, consistently favoring \MethodName. 
For Action \& Motion Alignment (FG \& BG), \MethodName\ achieves notably high win rates, highlighting a key limitation of I2V baselines that discard temporal dynamics by operating on static frames. 
In contrast, conditioning on full video layers enables \MethodName\ to preserve motion. 
It is also strongly preferred for FG-BG Harmony (often $>$85\%), demonstrating effective affordance-aware integration. 
High Overall Quality scores further reflect these advantages. 
Although Qwen + I2V is competitive in perceived quality, it performs poorly on motion and identity metrics, indicating that visual appeal alone does not ensure dynamic or identity fidelity. Appendix~\ref{sec:affordance_eval} provides a focused evaluation with a greater emphasis on affordance awareness.

\begin{figure*}[t!]
    \centering
    \includegraphics[width=\textwidth]{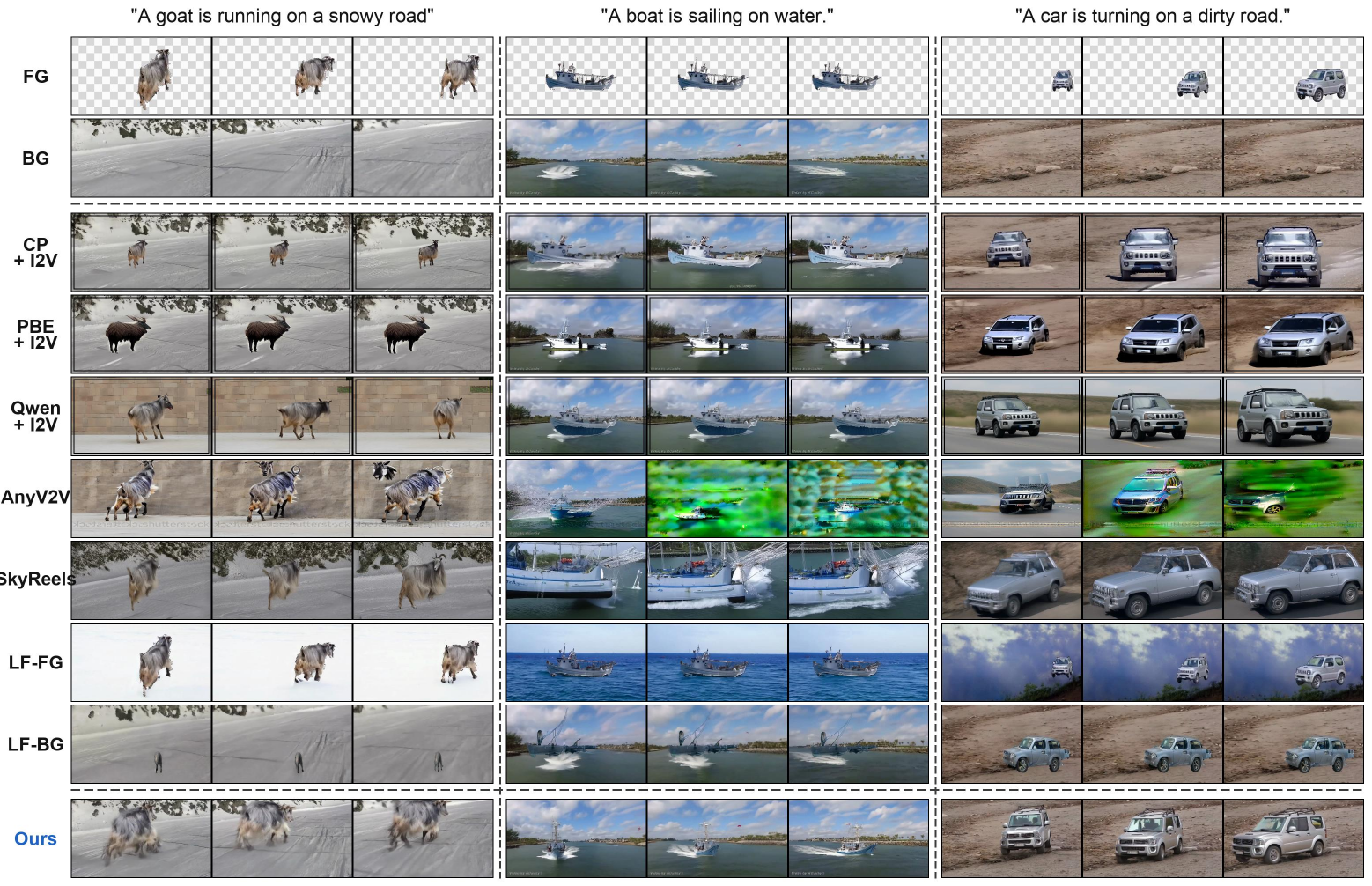}
    \vspace{-5ex}
    \caption{{\textbf{Qualitative comparison.}} 
    Our method (\MethodName) uniquely preserves complex dynamics and achieves affordance-aware harmony where baselines fail. 
    \textbf{(Left)} \MethodName\ alone maintains both the rapid background camera motion and realistic foreground running motion. 
    \textbf{(Center)} \MethodName\ demonstrates affordance by adapting the boat's orientation and height of the waves. 
    \textbf{(Right)} \MethodName\ accurately preserves the car's semantic action, road alignment, and lighting consistency, unlike alternative methods. Full videos are available on our \href{https://split-then-merge.github.io}{project page}.
    }
    \label{fig:qual-comp}
    \vspace{1ex}
\end{figure*}

\paragraph{Qualitative Evaluation.}
Fig.~\ref{fig:qual-res} and~\ref{fig:qual-comp} demonstrate \MethodName's ability to produce affordance-aware, dynamically coherent compositions. 
As shown in Fig.~\ref{fig:qual-comp}, \MethodName\ preserves rapid camera motion with a running goat (Left), correctly orients and positions a boat with wave dynamics (Center), and maintains a car's complex turn while aligning it with road geometry and lighting (Right). 
Fig.~\ref{fig:qual-res} further illustrates robust identity and motion preservation (cheetah, tennis player), alongside realistic integration of subjects (swan, duck, lion) with water flow, shadows, and illumination, avoiding a pasted appearance.

\paragraph{Ablation Study.}
Table~\ref{tab:quant_eval_and_ablation} presents ablation results. 
The \textit{w/o Both} variant learns a harmful copy-and-paste shortcut, yielding inflated FG scores (90.39 FG Identity, 0.77 FG Action) but poor compositional quality (18.59 BG Motion). 
Adding transformation-aware augmentation disrupts this shortcut: without \textit{ID Loss}, FG Identity drops to 82.01, and FG Action slightly increases to 0.92, reflecting genuine composition. 
Removing \textit{Augmentation} alone fails to prevent shortcut learning (0.70 FG Action). 
The full \textbf{\MethodName} balances both components, and achieves lower FG scores (84.82 FG Identity, 1.22 FG Action) but better BG Identity (92.88) and BG Motion (16.36).
This confirms the need for both modules in robust, affordance-aware composition.

\vspace{-1ex}
\subsection{In-the-Wild Evaluation}
\vspace{-1ex}

\label{sec:applications}
We further evaluate \MethodName\ through a series of \textit{stress tests} along three dimensions: (i) generalization to real-world background videos with diverse scenes and lighting; (ii) logically inconsistent composition prompts; and (iii) iterative multi-object insertion. 
Foregrounds are drawn from the \MethodName\ test set, while backgrounds are handheld videos captured in varied environments (\eg, parks, backyards, living rooms) under changing illumination.

\begin{figure*}[t!]
\centering
\resizebox{\textwidth}{!}{%
\includegraphics{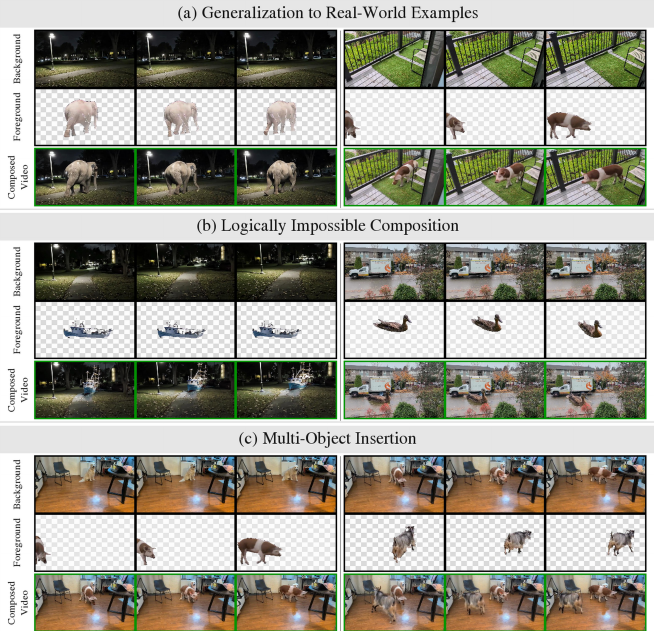}%
}
     \caption{{{\textbf{StM In-the-Wild.}} \MethodName~is stress-tested to show its ability to: 
     (a) generalize to smartphone videos with varying scenes and lighting,
     (b) creatively resolve logically impossible prompts such as ``a boat floating on a road'',
     and (c) seamlessly integrate multiple objects into a single background via recursive composition. 
     Full-resolution videos and extended results are available on our \href{https://split-then-merge.github.io}{project page}. 
     }}
    \label{fig:qual-results-app}
\end{figure*}

\paragraph{Generalization to Real-World Examples.}
Fig.~\ref{fig:qual-results-app}(a) shows robust performance across scenes and lighting. 
\MethodName\ seamlessly inserts an elephant into a night-time park, adapting illumination to ambient conditions, and reposes a pig to harmonize with balcony geometry and context.

\paragraph{Logically Impossible Composition.}
Fig.~\ref{fig:qual-results-app}(b) presents results under semantically inconsistent prompts. 
\MethodName\ re-orients a boat to align with road perspective while synthesizing water on asphalt, and transforms part of a road into water to accommodate a swimming duck. 
These examples highlight strong geometric reasoning and affordance awareness.

\paragraph{Multi-Object Insertion.}
Fig.~\ref{fig:qual-results-app}(c) demonstrates iterative composition, inserting a pig and a goat sequentially into a living room. 
\MethodName\ grounds both subjects on the floor and synthesizes consistent shadows, indicating scalability to complex multi-object scenarios.

\begin{figure*}[t!]
\centering
\resizebox{\textwidth}{!}{%
\includegraphics{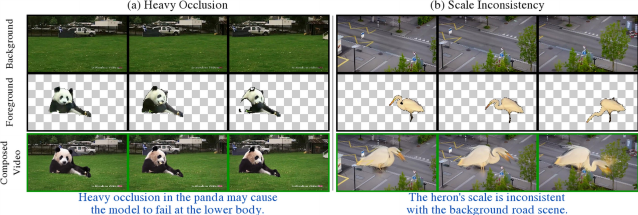}%
}
    \caption{{{\textbf{Challenging Test Cases}. 
    (a) Severe occlusion in the foreground prevents full reconstruction of the panda. 
    (b) Scale ambiguity from StM placing the heron with disproportionate size on the street. 
    Videos are available on our \href{https://split-then-merge.github.io}{project page}.}}}
    \label{fig:qual-failure}
\end{figure*}

\paragraph{Challenging Test Cases.} Fig.~\ref{fig:qual-failure} illustrates challenging test cases for \MethodName~caused by severe occlusion and scale ambiguity. In the first example (Fig.~\ref{fig:qual-failure}(a)), \MethodName~cannot reconstruct the entire panda due to severe occlusion in the foreground input. However, it still manages to partially recover the subject and harmonize its colors with the background scene. In the second example (Fig.~\ref{fig:qual-failure}(b)), although \MethodName~plausibly grounds the heron on the street, the subject's scale is disproportionately large relative to the rest of the scene.

\vspace{-1ex}

\subsection{Robustness to Decomposer Quality}
\label{sec:decomposer_robustness}
Because \MethodName\ relies on an off-the-shelf Decomposer to produce its foreground (FG) layers, we evaluate its robustness to imperfect FG masks. We corrupt the FG masks during inference via two perturbations of increasing severity: \emph{random pixel masking} (randomly removing interior pixels of the FG mask) and \emph{boundary erosion} (removing pixels along the FG boundary). We then evaluate \MethodName\ composition under these corrupted FGs (Table~\ref{tab:noise}).

\begin{figure}[t!]
    \centering
    \includegraphics[width=\linewidth]{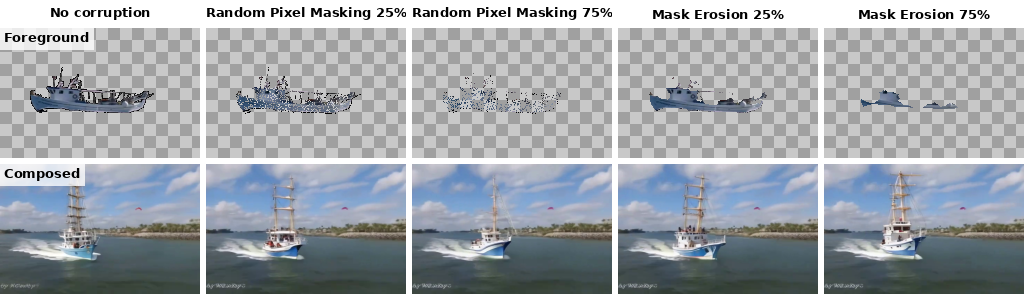}
    \caption{\textbf{\MethodName\ composition under corrupted foregrounds.} Randomly-masked and eroded foreground regions are shown for several severity levels; \MethodName\ recovers a coherent composition in each case.}
    \label{fig:corruption}
\end{figure}

\MethodName\ degrades gracefully, consistently outperforming LayerFlow-FG/BG across all metrics (M1--M4) and severities. It also surpasses AnyV2V entirely on M2 and M4, and maintains its lead on M1 (up to 25\% masking and 50\% erosion) and M3 (up to 10\% masking and 25\% erosion). This robustness is due to: (i) supervised losses operating in a latent space, making \MethodName\ less sensitive to pixel variations; (ii) a highly diverse training dataset; and (iii) the use of imperfect FG inputs during training. Fig.~\ref{fig:corruption} shows \MethodName\ composed results under different levels and types of corrupted FG.

Besides affecting inference, the Decomposer may also bias the evaluation metrics (M1--M4). We provide a full comparison of \MethodName\ with other baselines using three different Decomposer options in Appendix~\ref{sec:decomposer_independent}. The main findings still hold and are independent of the choice of Decomposer.

\begin{table}[t!]
  \centering
  \setlength{\tabcolsep}{6pt}
  \renewcommand{\arraystretch}{0.9}
  \caption{\textbf{Decomposer corruption} (evaluated with SAM-3). \MethodName\ composition under random pixel masking and boundary erosion of the foreground masks, compared against the clean setting and the video baselines.}
  \vspace{-1.5ex}
  \label{tab:noise}
  \begin{tabular}{lccccc}
  \toprule
   & {\bf M1}$\uparrow$ & {\bf M2}$\uparrow$ & {\bf M3}$\downarrow$ & {\bf M4}$\downarrow$ & {\bf M5}$\uparrow$\\
  \midrule
  \textit{Clean}        & 82.22 & 90.60 & 1.06 &  16.88 & 19.81 \\
  \midrule
  \textit{LayerFlow-FG} & 53.99 & 26.35 & 3.97 & 176.24 & 18.74 \\
  \textit{LayerFlow-BG} & 51.93 & 80.15 & 4.12 &  79.13 & 19.57 \\
  \textit{AnyV2V}       & 75.90 & 56.58 & 1.37 & 175.36 & 24.13 \\
  \midrule
  Rand.\ Pixel Masking 10\%   & 79.59 & 89.83 & 1.29 & 17.38 & 19.26 \\
  Rand.\ Pixel Masking 25\%   & 77.08 & 88.66 & 1.41 & 17.33 & 19.29 \\
  Rand.\ Pixel Masking 50\%   & 73.27 & 87.07 & 1.72 & 26.94 & 18.47 \\
  Rand.\ Pixel Masking 75\%   & 73.87 & 88.22 & 1.81 & 27.00 & 18.47 \\
  \midrule
  Mask Erosion 10\%        & 79.09 & 89.08 & 1.30 & 17.20 & 19.42 \\
  Mask Erosion 25\%        & 78.28 & 88.49 & 1.32 & 17.24 & 19.37 \\
  Mask Erosion 50\%        & 76.86 & 87.40 & 1.44 & 19.35 & 19.11 \\
  Mask Erosion 75\%        & 74.06 & 87.51 & 1.63 & 21.20 & 19.03 \\
  \bottomrule
  \end{tabular}
\end{table}

\subsection{Limitation}
\label{sec:limitation}
\vspace{-1ex}

\MethodName~still occasionally generates videos with scale inconsistencies between foreground objects and the background scene, \eg, Fig.~\ref{fig:qual-failure}(b), the duck example in Fig.~\ref{fig:qual-results-app}(b), and Fig.~\ref{fig:motivation}(e). It is inherently challenging for a purely data-driven approach to automatically deduce the relative sizes of disparate semantic categories---such as understanding that a duck is similar in size to a swan, but significantly smaller than a boat. Future work will explore incorporating relative scale priors for semantic objects and scenes directly into the training pipeline.

\vspace{-1ex}
\section{Conclusion}
\vspace{-1ex}

We present \MethodName, a practical pipeline for generative video composition that addresses data scarcity via a scalable ``Split-then-Merge'' strategy. 
By decomposing unlabeled videos into dynamic layers and self-composing them, \MethodName\ learns subject-scene composition without manual annotation. 
Built on the \MethodName-50K multi-layer dataset, our approach incorporates two key components: an \textit{identity-preservation loss} for subject fidelity and \textit{transformation-aware training} for motion and affordance modeling. 
Extensive experiments demonstrate that \MethodName\ outperforms prior methods in motion preservation and compositional harmony.

\newpage

\bibliographystyle{splncs04}
\bibliography{main}

@String(ICCV= {Int. Conf. Comput. Vis.})

@String(AAAI = {AAAI})

@String(ICCV  = {ICCV})

@article{ho2020denoising,
  title={Denoising diffusion probabilistic models},
  author={Ho, Jonathan and Jain, Ajay and Abbeel, Pieter},
  journal={Advances in neural information processing systems},
  volume={33},
  pages={6840--6851},
  year={2020}
}

@inproceedings{rombach2022high,
  title={High-resolution image synthesis with latent diffusion models},
  author={Rombach, Robin and Blattmann, Andreas and Lorenz, Dominik and Esser, Patrick and Ommer, Bj{\"o}rn},
  booktitle={Proceedings of the IEEE/CVF conference on computer vision and pattern recognition},
  pages={10684--10695},
  year={2022}
}

@inproceedings{yang2023paint,
  title={Paint by example: Exemplar-based image editing with diffusion models},
  author={Yang, Binxin and Gu, Shuyang and Zhang, Bo and Zhang, Ting and Chen, Xuejin and Sun, Xiaoyan and Chen, Dong and Wen, Fang},
  booktitle={Proceedings of the IEEE/CVF conference on computer vision and pattern recognition},
  pages={18381--18391},
  year={2023}
}

@inproceedings{chen2024anydoor,
  title={Anydoor: Zero-shot object-level image customization},
  author={Chen, Xi and Huang, Lianghua and Liu, Yu and Shen, Yujun and Zhao, Deli and Zhao, Hengshuang},
  booktitle={Proceedings of the IEEE/CVF conference on computer vision and pattern recognition},
  pages={6593--6602},
  year={2024}
}

@inproceedings{
yang2025cogvideox,
title={CogVideoX: Text-to-Video Diffusion Models with An Expert Transformer},
author={Zhuoyi Yang and Jiayan Teng and Wendi Zheng and Ming Ding and Shiyu Huang and Jiazheng Xu and Yuanming Yang and Wenyi Hong and Xiaohan Zhang and Guanyu Feng and Da Yin and Yuxuan.Zhang and Weihan Wang and Yean Cheng and Bin Xu and Xiaotao Gu and Yuxiao Dong and Jie Tang},
booktitle={The Thirteenth International Conference on Learning Representations},
year={2025},
url={https://openreview.net/forum?id=LQzN6TRFg9}
}

@inproceedings{peebles2023scalable,
  title={Scalable diffusion models with transformers},
  author={Peebles, William and Xie, Saining},
  booktitle={Proceedings of the IEEE/CVF international conference on computer vision},
  pages={4195--4205},
  year={2023}
}

@inproceedings{chen2024internvl,
  title={Internvl: Scaling up vision foundation models and aligning for generic visual-linguistic tasks},
  author={Chen, Zhe and Wu, Jiannan and Wang, Wenhai and Su, Weijie and Chen, Guo and Xing, Sen and Zhong, Muyan and Zhang, Qinglong and Zhu, Xizhou and Lu, Lewei and others},
  booktitle={Proceedings of the IEEE/CVF conference on computer vision and pattern recognition},
  pages={24185--24198},
  year={2024}
}

@inproceedings{huang2025segment,
  title={Segment Any Motion in Videos},
  author={Huang, Nan and Zheng, Wenzhao and Xu, Chenfeng and Keutzer, Kurt and Zhang, Shanghang and Kanazawa, Angjoo and Wang, Qianqian},
  booktitle={Proceedings of the Computer Vision and Pattern Recognition Conference},
  pages={3406--3416},
  year={2025}
}

@inproceedings{perazzi2016benchmark,
  title={A benchmark dataset and evaluation methodology for video object segmentation},
  author={Perazzi, Federico and Pont-Tuset, Jordi and McWilliams, Brian and Van Gool, Luc and Gross, Markus and Sorkine-Hornung, Alexander},
  booktitle={Proceedings of the IEEE conference on computer vision and pattern recognition},
  pages={724--732},
  year={2016}
}

@article{xu2018youtube,
  title={Youtube-vos: A large-scale video object segmentation benchmark},
  author={Xu, Ning and Yang, Linjie and Fan, Yuchen and Yue, Dingcheng and Liang, Yuchen and Yang, Jianchao and Huang, Thomas},
  journal={arXiv preprint arXiv:1809.03327},
  year={2018}
}

@inproceedings{hong2023lvos,
  title={Lvos: A benchmark for long-term video object segmentation},
  author={Hong, Lingyi and Chen, Wenchao and Liu, Zhongying and Zhang, Wei and Guo, Pinxue and Chen, Zhaoyu and Zhang, Wenqiang},
  booktitle={Proceedings of the IEEE/CVF International Conference on Computer Vision},
  pages={13480--13492},
  year={2023}
}

@inproceedings{chen2024panda,
  title={Panda-70m: Captioning 70m videos with multiple cross-modality teachers},
  author={Chen, Tsai-Shien and Siarohin, Aliaksandr and Menapace, Willi and Deyneka, Ekaterina and Chao, Hsiang-wei and Jeon, Byung Eun and Fang, Yuwei and Lee, Hsin-Ying and Ren, Jian and Yang, Ming-Hsuan and others},
  booktitle={Proceedings of the IEEE/CVF Conference on Computer Vision and Pattern Recognition},
  pages={13320--13331},
  year={2024}
}

@inproceedings{ng2022animal,
  title={Animal kingdom: A large and diverse dataset for animal behavior understanding},
  author={Ng, Xun Long and Ong, Kian Eng and Zheng, Qichen and Ni, Yun and Yeo, Si Yong and Liu, Jun},
  booktitle={Proceedings of the IEEE/CVF conference on computer vision and pattern recognition},
  pages={19023--19034},
  year={2022}
}

@inproceedings{tu2025videoanydoor,
  title={Videoanydoor: High-fidelity video object insertion with precise motion control},
  author={Tu, Yuanpeng and Luo, Hao and Chen, Xi and Ji, Sihui and Bai, Xiang and Zhao, Hengshuang},
  booktitle={Proceedings of the Special Interest Group on Computer Graphics and Interactive Techniques Conference Conference Papers},
  pages={1--11},
  year={2025}
}

@article{zi2025minimax,
  title={MiniMax-Remover: Taming Bad Noise Helps Video Object Removal},
  author={Zi, Bojia and Peng, Weixuan and Qi, Xianbiao and Wang, Jianan and Zhao, Shihao and Xiao, Rong and Wong, Kam-Fai},
  journal={arXiv preprint arXiv:2505.24873},
  year={2025}
}

@inproceedings{
song2021denoising,
title={Denoising Diffusion Implicit Models},
author={Jiaming Song and Chenlin Meng and Stefano Ermon},
booktitle={International Conference on Learning Representations},
year={2021},
url={https://openreview.net/forum?id=St1giarCHLP}
}

@inproceedings{guo2024i2v,
  title={I2v-adapter: A general image-to-video adapter for diffusion models},
  author={Guo, Xun and Zheng, Mingwu and Hou, Liang and Gao, Yuan and Deng, Yufan and Wan, Pengfei and Zhang, Di and Liu, Yufan and Hu, Weiming and Zha, Zhengjun and others},
  booktitle={ACM SIGGRAPH 2024 Conference Papers},
  pages={1--12},
  year={2024}
}

@article{
ren2024consistiv,
title={ConsistI2V: Enhancing Visual Consistency for Image-to-Video Generation},
author={Weiming Ren and Huan Yang and Ge Zhang and Cong Wei and Xinrun Du and Wenhao Huang and Wenhu Chen},
journal={Transactions on Machine Learning Research},
issn={2835-8856},
year={2024},
url={https://openreview.net/forum?id=vqniLmUDvj},
note={}
}

@inproceedings{shi2024motion,
  title={Motion-i2v: Consistent and controllable image-to-video generation with explicit motion modeling},
  author={Shi, Xiaoyu and Huang, Zhaoyang and Wang, Fu-Yun and Bian, Weikang and Li, Dasong and Zhang, Yi and Zhang, Manyuan and Cheung, Ka Chun and See, Simon and Qin, Hongwei and others},
  booktitle={ACM SIGGRAPH 2024 Conference Papers},
  pages={1--11},
  year={2024}
}

@inproceedings{kara2024rave,
  title={Rave: Randomized noise shuffling for fast and consistent video editing with diffusion models},
  author={Kara, Ozgur and Kurtkaya, Bariscan and Yesiltepe, Hidir and Rehg, James M and Yanardag, Pinar},
  booktitle={Proceedings of the IEEE/CVF Conference on Computer Vision and Pattern Recognition},
  pages={6507--6516},
  year={2024}
}

@inproceedings{
geyer2024tokenflow,
title={TokenFlow: Consistent Diffusion Features for Consistent Video Editing},
author={Michal Geyer and Omer Bar-Tal and Shai Bagon and Tali Dekel},
booktitle={The Twelfth International Conference on Learning Representations},
year={2024},
url={https://openreview.net/forum?id=lKK50q2MtV}
}

@inproceedings{
namekata2025sgiv,
title={{SG}-I2V: Self-Guided Trajectory Control in Image-to-Video Generation},
author={Koichi Namekata and Sherwin Bahmani and Ziyi Wu and Yash Kant and Igor Gilitschenski and David B. Lindell},
booktitle={The Thirteenth International Conference on Learning Representations},
year={2025},
url={https://openreview.net/forum?id=uQjySppU9x}
}

@inproceedings{ouyang2024i2vedit,
  title={I2vedit: First-frame-guided video editing via image-to-video diffusion models},
  author={Ouyang, Wenqi and Dong, Yi and Yang, Lei and Si, Jianlou and Pan, Xingang},
  booktitle={SIGGRAPH Asia 2024 Conference Papers},
  pages={1--11},
  year={2024}
}

@inproceedings{
feng2025ivcontrolcamera,
title={I2{VC}ontrol-Camera: Precise Video Camera Control with Adjustable Motion Strength},
author={Wanquan Feng and Jiawei Liu and Pengqi Tu and Tianhao Qi and Mingzhen Sun and Tianxiang Ma and Songtao Zhao and SiYu Zhou and Qian HE},
booktitle={The Thirteenth International Conference on Learning Representations},
year={2025},
url={https://openreview.net/forum?id=AcAD4VEgCX}
}

@inproceedings{
hong2023cogvideo,
title={CogVideo: Large-scale Pretraining for Text-to-Video Generation via Transformers},
author={Wenyi Hong and Ming Ding and Wendi Zheng and Xinghan Liu and Jie Tang},
booktitle={The Eleventh International Conference on Learning Representations },
year={2023},
url={https://openreview.net/forum?id=rB6TpjAuSRy}
}

@inproceedings{bar2024lumiere,
  title={Lumiere: A space-time diffusion model for video generation},
  author={Bar-Tal, Omer and Chefer, Hila and Tov, Omer and Herrmann, Charles and Paiss, Roni and Zada, Shiran and Ephrat, Ariel and Hur, Junhwa and Liu, Guanghui and Raj, Amit and others},
  booktitle={SIGGRAPH Asia 2024 Conference Papers},
  pages={1--11},
  year={2024}
}

@inproceedings{gupta2024photorealistic,
  title={Photorealistic video generation with diffusion models},
  author={Gupta, Agrim and Yu, Lijun and Sohn, Kihyuk and Gu, Xiuye and Hahn, Meera and Li, Fei-Fei and Essa, Irfan and Jiang, Lu and Lezama, Jos{\'e}},
  booktitle={European Conference on Computer Vision},
  pages={393--411},
  year={2024},
  organization={Springer}
}

@inproceedings{ma2024follow,
  title={Follow your pose: Pose-guided text-to-video generation using pose-free videos},
  author={Ma, Yue and He, Yingqing and Cun, Xiaodong and Wang, Xintao and Chen, Siran and Li, Xiu and Chen, Qifeng},
  booktitle={Proceedings of the AAAI Conference on Artificial Intelligence},
  volume={38},
  number={5},
  pages={4117--4125},
  year={2024}
}

@inproceedings{
chang2024magicpose,
title={MagicPose: Realistic Human Poses and Facial Expressions Retargeting with Identity-aware Diffusion},
author={Di Chang and Yichun Shi and Quankai Gao and Hongyi Xu and Jessica Fu and Guoxian Song and Qing Yan and Yizhe Zhu and Xiao Yang and Mohammad Soleymani},
booktitle={Forty-first International Conference on Machine Learning},
year={2024},
url={https://openreview.net/forum?id=jVXJdGQ4eD}
}

@inproceedings{karras2023dreampose,
  title={Dreampose: Fashion video synthesis with stable diffusion},
  author={Karras, Johanna and Holynski, Aleksander and Wang, Ting-Chun and Kemelmacher-Shlizerman, Ira},
  booktitle={Proceedings of the IEEE/CVF International Conference on Computer Vision},
  pages={22680--22690},
  year={2023}
}

@article{mou2024revideo,
  title={Revideo: Remake a video with motion and content control},
  author={Mou, Chong and Cao, Mingdeng and Wang, Xintao and Zhang, Zhaoyang and Shan, Ying and Zhang, Jian},
  journal={Advances in Neural Information Processing Systems},
  volume={37},
  pages={18481--18505},
  year={2024}
}

@inproceedings{hu2024animate,
  title={Animate anyone: Consistent and controllable image-to-video synthesis for character animation},
  author={Hu, Li},
  booktitle={Proceedings of the IEEE/CVF Conference on Computer Vision and Pattern Recognition},
  pages={8153--8163},
  year={2024}
}

@inproceedings{zhang2023adding,
  title={Adding conditional control to text-to-image diffusion models},
  author={Zhang, Lvmin and Rao, Anyi and Agrawala, Maneesh},
  booktitle={Proceedings of the IEEE/CVF international conference on computer vision},
  pages={3836--3847},
  year={2023}
}

@inproceedings{geng2025motion,
  title={Motion prompting: Controlling video generation with motion trajectories},
  author={Geng, Daniel and Herrmann, Charles and Hur, Junhwa and Cole, Forrester and Zhang, Serena and Pfaff, Tobias and Lopez-Guevara, Tatiana and Aytar, Yusuf and Rubinstein, Michael and Sun, Chen and others},
  booktitle={Proceedings of the Computer Vision and Pattern Recognition Conference},
  pages={1--12},
  year={2025}
}

@inproceedings{burgert2025go,
  title={Go-with-the-flow: Motion-controllable video diffusion models using real-time warped noise},
  author={Burgert, Ryan and Xu, Yuancheng and Xian, Wenqi and Pilarski, Oliver and Clausen, Pascal and He, Mingming and Ma, Li and Deng, Yitong and Li, Lingxiao and Mousavi, Mohsen and others},
  booktitle={Proceedings of the Computer Vision and Pattern Recognition Conference},
  pages={13--23},
  year={2025}
}

@book{wright2013digital,
  title={Digital compositing for film and video},
  author={Wright, Steve},
  year={2013},
  publisher={Routledge}
}

@book{brinkmann2008art,
  title={The art and science of digital compositing: Techniques for visual effects, animation and motion graphics},
  author={Brinkmann, Ron},
  year={2008},
  publisher={Morgan Kaufmann}
}

@inproceedings{ji2025layerflow,
  title={LayerFlow: A Unified Model for Layer-aware Video Generation},
  author={Ji, Sihui and Luo, Hao and Chen, Xi and Tu, Yuanpeng and Wang, Yiyang and Zhao, Hengshuang},
  booktitle={Proceedings of the Special Interest Group on Computer Graphics and Interactive Techniques Conference Conference Papers},
  pages={1--10},
  year={2025}
}

@article{
ku2024anyvv,
title={AnyV2V: A Tuning-Free Framework For Any Video-to-Video Editing Tasks},
author={Max Ku and Cong Wei and Weiming Ren and Huan Yang and Wenhu Chen},
journal={Transactions on Machine Learning Research},
issn={2835-8856},
year={2024},
url={https://openreview.net/forum?id=RFrJCkw2oa},
note={Reproducibility Certification}
}

@inproceedings{khachatryan2023text2video,
  title={Text2video-zero: Text-to-image diffusion models are zero-shot video generators},
  author={Khachatryan, Levon and Movsisyan, Andranik and Tadevosyan, Vahram and Henschel, Roberto and Wang, Zhangyang and Navasardyan, Shant and Shi, Humphrey},
  booktitle={Proceedings of the IEEE/CVF International Conference on Computer Vision},
  pages={15954--15964},
  year={2023}
}

@inproceedings{zhao2024magdiff,
  title={Magdiff: Multi-alignment diffusion for high-fidelity video generation and editing},
  author={Zhao, Haoyu and Lu, Tianyi and Gu, Jiaxi and Zhang, Xing and Zheng, Qingping and Wu, Zuxuan and Xu, Hang and Jiang, Yu-Gang},
  booktitle={European Conference on Computer Vision},
  pages={205--221},
  year={2024},
  organization={Springer}
}

@inproceedings{jiang2024videobooth,
  title={Videobooth: Diffusion-based video generation with image prompts},
  author={Jiang, Yuming and Wu, Tianxing and Yang, Shuai and Si, Chenyang and Lin, Dahua and Qiao, Yu and Loy, Chen Change and Liu, Ziwei},
  booktitle={Proceedings of the IEEE/CVF Conference on Computer Vision and Pattern Recognition},
  pages={6689--6700},
  year={2024}
}

@inproceedings{guo2024sparsectrl,
  title={Sparsectrl: Adding sparse controls to text-to-video diffusion models},
  author={Guo, Yuwei and Yang, Ceyuan and Rao, Anyi and Agrawala, Maneesh and Lin, Dahua and Dai, Bo},
  booktitle={European Conference on Computer Vision},
  pages={330--348},
  year={2024},
  organization={Springer}
}

@InProceedings{pmlr-v235-wang24cr,
  title = 	 {Boximator: Generating Rich and Controllable Motions for Video Synthesis},
  author =       {Wang, Jiawei and Zhang, Yuchen and Zou, Jiaxin and Zeng, Yan and Wei, Guoqiang and Yuan, Liping and Li, Hang},
  booktitle = 	 {Proceedings of the 41st International Conference on Machine Learning},
  pages = 	 {52274--52289},
  year = 	 {2024},
  editor = 	 {Salakhutdinov, Ruslan and Kolter, Zico and Heller, Katherine and Weller, Adrian and Oliver, Nuria and Scarlett, Jonathan and Berkenkamp, Felix},
  volume = 	 {235},
  series = 	 {Proceedings of Machine Learning Research},
  month = 	 {21--27 Jul},
  publisher =    {PMLR},
  url = 	 {https://proceedings.mlr.press/v235/wang24cr.html}
}

@article{wu2024motionbooth,
  title={Motionbooth: Motion-aware customized text-to-video generation},
  author={Wu, Jianzong and Li, Xiangtai and Zeng, Yanhong and Zhang, Jiangning and Zhou, Qianyu and Li, Yining and Tong, Yunhai and Chen, Kai},
  journal={Advances in Neural Information Processing Systems},
  volume={37},
  pages={34322--34348},
  year={2024}
}

@inproceedings{wu2024draganything,
  title={Draganything: Motion control for anything using entity representation},
  author={Wu, Weijia and Li, Zhuang and Gu, Yuchao and Zhao, Rui and He, Yefei and Zhang, David Junhao and Shou, Mike Zheng and Li, Yan and Gao, Tingting and Zhang, Di},
  booktitle={European Conference on Computer Vision},
  pages={331--348},
  year={2024},
  organization={Springer}
}

@inproceedings{
zhang2024controlvideo,
title={ControlVideo: Training-free Controllable Text-to-video Generation},
author={Yabo Zhang and Yuxiang Wei and Dongsheng Jiang and XIAOPENG ZHANG and Wangmeng Zuo and Qi Tian},
booktitle={The Twelfth International Conference on Learning Representations},
year={2024},
url={https://openreview.net/forum?id=5a79AqFr0c}
}

@misc{wu2025qwenimagetechnicalreport,
      title={Qwen-Image Technical Report}, 
      author={Chenfei Wu and Jiahao Li and Jingren Zhou and Junyang Lin and Kaiyuan Gao and Kun Yan and Sheng-ming Yin and Shuai Bai and Xiao Xu and Yilei Chen and Yuxiang Chen and Zecheng Tang and Zekai Zhang and Zhengyi Wang and An Yang and Bowen Yu and Chen Cheng and Dayiheng Liu and Deqing Li and Hang Zhang and Hao Meng and Hu Wei and Jingyuan Ni and Kai Chen and Kuan Cao and Liang Peng and Lin Qu and Minggang Wu and Peng Wang and Shuting Yu and Tingkun Wen and Wensen Feng and Xiaoxiao Xu and Yi Wang and Yichang Zhang and Yongqiang Zhu and Yujia Wu and Yuxuan Cai and Zenan Liu},
      year={2025},
      eprint={2508.02324},
      archivePrefix={arXiv},
      primaryClass={cs.CV},
      url={https://arxiv.org/abs/2508.02324}, 
}

@ARTICLE{Chen2013Motion,
author={Tao Chen and Jun-Yan Zhu and Shamir, A. and Shi-Min Hu},
journal={Image Processing, IEEE Transactions on},
title={Motion-Aware Gradient Domain Video Composition},
year={2013},
volume={22},
number={7},
pages={2532-2544},
doi={10.1109/TIP.2013.2251642},
ISSN={1057-7149},
}

@article{Wang2019Illumination,
author = {Wang, Jingye and Sheng, Bin and Li, Ping and Jin, Yuxi and Feng, David Dagan},
title = {Illumination-Guided Video Composition via Gradient Consistency Optimization},
year = {2019},
issue_date = {Oct. 2019},
publisher = {IEEE Press},
volume = {28},
number = {10},
issn = {1057-7149},
url = {https://doi.org/10.1109/TIP.2019.2916769},
doi = {10.1109/TIP.2019.2916769},
journal = {Trans. Img. Proc.},
month = oct,
pages = {5077–5090},
numpages = {14}
}

@inproceedings{chen2025videoalchemist,
  title   = {Multi-subject Open-set Personalization in Video Generation},
  author  = {Chen, Tsai-Shien and Siarohin, Aliaksandr and Menapace, Willi and Fang, Yuwei and Lee, Kwot Sin and Skorokhodov, Ivan and Aberman, Kfir and Zhu, Jun-Yan and Yang, Ming-Hsuan and Tulyakov, Sergey},
  journal = {Proceedings of the IEEE/CVF Conference on Computer Vision and Pattern Recognition},
  year    = {2025}
}

@article{fei2025skyreels,
  title={SkyReels-A2: Compose Anything in Video Diffusion Transformers},
  author={Fei, Zhengcong and Li, Debang and Qiu, Di and Wang, Jiahua and Dou, Yikun and Wang, Rui and Xu, Jingtao and Fan, Mingyuan and Chen, Guibin and Li, Yang and others},
  journal={arXiv preprint arXiv:2504.02436},
  year={2025}
}

@inproceedings{pitie2005n,
  title={N-dimensional probability density function transfer and its application to color transfer},
  author={Pitie, F. and Kokaram, A. C. and Dahyot, R.},
  booktitle={ICCV},
  year={2005}
}

@article{reinhard2001color,
  title={Color transfer between images},
  author={Reinhard, Erik and Adhikhmin, Michael and Gooch, Bruce and Shirley, Peter},
  journal={IEEE Computer graphics and applications},
  volume={21},
  number={5},
  pages={34--41},
  year={2001}
}

@article{jia2006drag,
  title={Drag-and-drop pasting},
  author={Jia, Jiaya and Sun, Jian and Tang, Chi-Keung and Shum, Heung-Yeung},
  journal={ACM Transactions on Graphics},
  volume={25},
  number={3},
  pages={631--637},
  year={2006}
}

@article{cohenor2006color,
  title={Color harmonization},
  author={Cohen-Or, Daniel and Sorkine, Olga and Gal, Ran and Leyvand, Tommer and Xu, Ying-Qing},
  journal={ACM Transactions on Graphics},
  volume={25},
  number={3},
  pages={624--630},
  year={2006}
}

@inproceedings{lalonde2007using,
  title={Using color compatibility for assessing image realism},
  author={Lalonde, Jean-Francois and Efros, Alexei A},
  booktitle={ICCV},
  year={2007}
}

@article{xue2012understanding,
  title={Understanding and improving the realism of image composites},
  author={Xue, Su and Agarwala, Aseem and Dorsey, Julie and Rushmeier, Holly},
  journal={ACM Transactions on Graphics},
  volume={31},
  number={4},
  pages={84},
  year={2012}
}

@inproceedings{
loshchilov2018decoupled,
title={Decoupled Weight Decay Regularization},
author={Ilya Loshchilov and Frank Hutter},
booktitle={International Conference on Learning Representations},
year={2019},
url={https://openreview.net/forum?id=Bkg6RiCqY7},
}

@inproceedings{
wang2024internvid,
title={InternVid: A Large-scale Video-Text Dataset for Multimodal Understanding and Generation},
author={Yi Wang and Yinan He and Yizhuo Li and Kunchang Li and Jiashuo Yu and Xin Ma and Xinhao Li and Guo Chen and Xinyuan Chen and Yaohui Wang and Ping Luo and Ziwei Liu and Yali Wang and Limin Wang and Yu Qiao},
booktitle={The Twelfth International Conference on Learning Representations},
year={2024},
url={https://openreview.net/forum?id=MLBdiWu4Fw}
}

@inproceedings{liu2022video,
  title={Video swin transformer},
  author={Liu, Ze and Ning, Jia and Cao, Yue and Wei, Yixuan and Zhang, Zheng and Lin, Stephen and Hu, Han},
  booktitle={Proceedings of the IEEE/CVF conference on computer vision and pattern recognition},
  pages={3202--3211},
  year={2022}
}

@inproceedings{
jaegle2022perceiver,
title={Perceiver {IO}: A General Architecture for Structured Inputs \& Outputs},
author={Andrew Jaegle and Sebastian Borgeaud and Jean-Baptiste Alayrac and Carl Doersch and Catalin Ionescu and David Ding and Skanda Koppula and Daniel Zoran and Andrew Brock and Evan Shelhamer and Olivier J Henaff and Matthew Botvinick and Andrew Zisserman and Oriol Vinyals and Joao Carreira},
booktitle={International Conference on Learning Representations},
year={2022},
url={https://openreview.net/forum?id=fILj7WpI-g}
}

@inproceedings{baliah2025realistic,
  title={Realistic and efficient face swapping: A unified approach with diffusion models},
  author={Baliah, Sanoojan and Lin, Qinliang and Liao, Shengcai and Liang, Xiaodan and Khan, Muhammad Haris},
  booktitle={2025 IEEE/CVF Winter Conference on Applications of Computer Vision (WACV)},
  pages={1062--1071},
  year={2025},
  organization={IEEE}
}

@inproceedings{luo2025canonswap,
  title={CanonSwap: High-Fidelity and Consistent Video Face Swapping via Canonical Space Modulation},
  author={Luo, Xiangyang and Zhu, Ye and Liu, Yunfei and Lin, Lijian and Wan, Cong and Cai, Zijian and Li, Yu and Huang, Shao-Lun},
  booktitle={Proceedings of the IEEE/CVF International Conference on Computer Vision},
  pages={10064--10074},
  year={2025}
}

@inproceedings{
shao2025vividface,
title={VividFace: A Robost and High-Fidelity Video Face Swapping Framework},
author={Hao Shao and Shulun Wang and Yang Zhou and Guanglu Song and Dailan He and Zhuofan Zong and Shuo Qin and Yu Liu and Hongsheng Li},
booktitle={The Thirty-ninth Annual Conference on Neural Information Processing Systems},
year={2025},
url={https://openreview.net/forum?id=wyv81ezGgv}
}

@article{pan2024actanywhere,
  title={Actanywhere: Subject-aware video background generation},
  author={Pan, Boxiao and Xu, Zhan and Huang, Chun-Hao and Singh, Krishna Kumar and Zhou, Yang and Guibas, Leonidas J and Yang, Jimei},
  journal={Advances in Neural Information Processing Systems},
  volume={37},
  pages={29754--29776},
  year={2024}
}

@inproceedings{gu2024videoswap,
  title={Videoswap: Customized video subject swapping with interactive semantic point correspondence},
  author={Gu, Yuchao and Zhou, Yipin and Wu, Bichen and Yu, Licheng and Liu, Jia-Wei and Zhao, Rui and Wu, Jay Zhangjie and Zhang, David Junhao and Shou, Mike Zheng and Tang, Kevin},
  booktitle={Proceedings of the IEEE/CVF Conference on Computer Vision and Pattern Recognition},
  pages={7621--7630},
  year={2024}
}

@article{comanici2025gemini,
  title={Gemini 2.5: Pushing the frontier with advanced reasoning, multimodality, long context, and next generation agentic capabilities},
  author={Comanici, Gheorghe and Bieber, Eric and Schaekermann, Mike and Pasupat, Ice and Sachdeva, Noveen and Dhillon, Inderjit and Blistein, Marcel and Ram, Ori and Zhang, Dan and Rosen, Evan and others},
  journal={arXiv preprint arXiv:2507.06261},
  year={2025}
}

@misc{prolific,
  author       = {Prolific},
  title        = {Prolific | Easily collect high-quality data from real people},
  year         = {2025},
  url          = {https://www.prolific.com/},
  urldate      = {2025-11-06}
}

\appendix
\clearpage
\setcounter{page}{1}
\begingroup
\renewcommand{\addcontentsline}[3]{}
\renewcommand{\addtocontents}[2]{}
\endgroup
\begingroup
\hypersetup{
    citecolor=eccvblue,
    linkcolor=eccvblue
}
\section*{Supplementary Contents}
\startcontents[appendices]
\printcontents[appendices]{}{1}{\setcounter{tocdepth}{2}}
\endgroup
\renewcommand{\thesection}{A\arabic{section}}
\renewcommand{\theHsection}{A\arabic{section}}
\section{StM-50K Dataset Details}
\label{sec:dataset_details}
\subsection{Data Sources and Composition}
\label{sec:data_sources}
The StM-50K multi-layer video dataset comprises approximately 50,000 video clips and was constructed using our automated Decomposer pipeline (Section~\ref{sec:decomposer}). For large, unannotated sources like Panda-70M~\cite{chen2024panda} and Animal Kingdom~\cite{ng2022animal}, the full Decomposer pipeline—including motion segmentation and video inpainting—was utilized to generate the foreground, mask, background, and caption layers. For existing video object segmentation (VOS) datasets (\eg, YouTube-VOS~\cite{xu2018youtube} and LVOS~\cite{hong2023lvos}), we adapted the pipeline to leverage their provided ground-truth foreground masks, thus ensuring high-fidelity layer separation. The DAVIS dataset~\cite{perazzi2016benchmark} was reserved exclusively for model validation during training.
To ensure temporal consistency and support multiple temporal resolution across the dataset, we apply a randomized preprocessing strategy. Videos are either directly split into multiple chunks of 49 frames, or first downsampled in frame rate (by a factor of 2, 3, or 4) and subsequently split into 49-frame chunks. This approach guarantees uniform input dimensions while preserving motion diversity.
\subsection{Effect of Data Scaling}
\label{sec:data_scaling}
To quantify how much of StM's performance is attributable to dataset size versus architectural design, we train the model on reduced subsets of StM-50K (Table~\ref{tab:datascale}). A 10K-trained StM ($5\times$ less data) still beats every baseline on M1--M4. The full 50K setting widens this margin, demonstrating that while data scale helps, the architecture is what matters most.
\begin{table}[t!]
\centering
\setlength{\tabcolsep}{3pt}
\renewcommand{\arraystretch}{0.9}
\caption{\textbf{Data scaling.} StM trained on increasing subsets of StM-50K. Even at $5\times$ less data (10K), StM outperforms all baselines on M1--M4.}
\label{tab:datascale}
\begin{tabular}{lccccc}
\toprule
 & M1$\uparrow$ & M2$\uparrow$ & M3$\downarrow$ & M4$\downarrow$ & M5$\uparrow$\\
\midrule
StM\ (10K)         & 77.96 & 88.11 & 1.64 & 22.27 & 17.63 \\
StM\ (20K)         & 81.18 & 91.23 & 1.60 & 20.85 & 18.26 \\
StM\ (50K, ours)   & 84.82 & 92.88 & 1.22 & 16.36 & 19.81 \\
\bottomrule
\end{tabular}
\end{table}
\begin{figure*}[t!]
    \centering
    \includegraphics[width=0.7\textwidth]{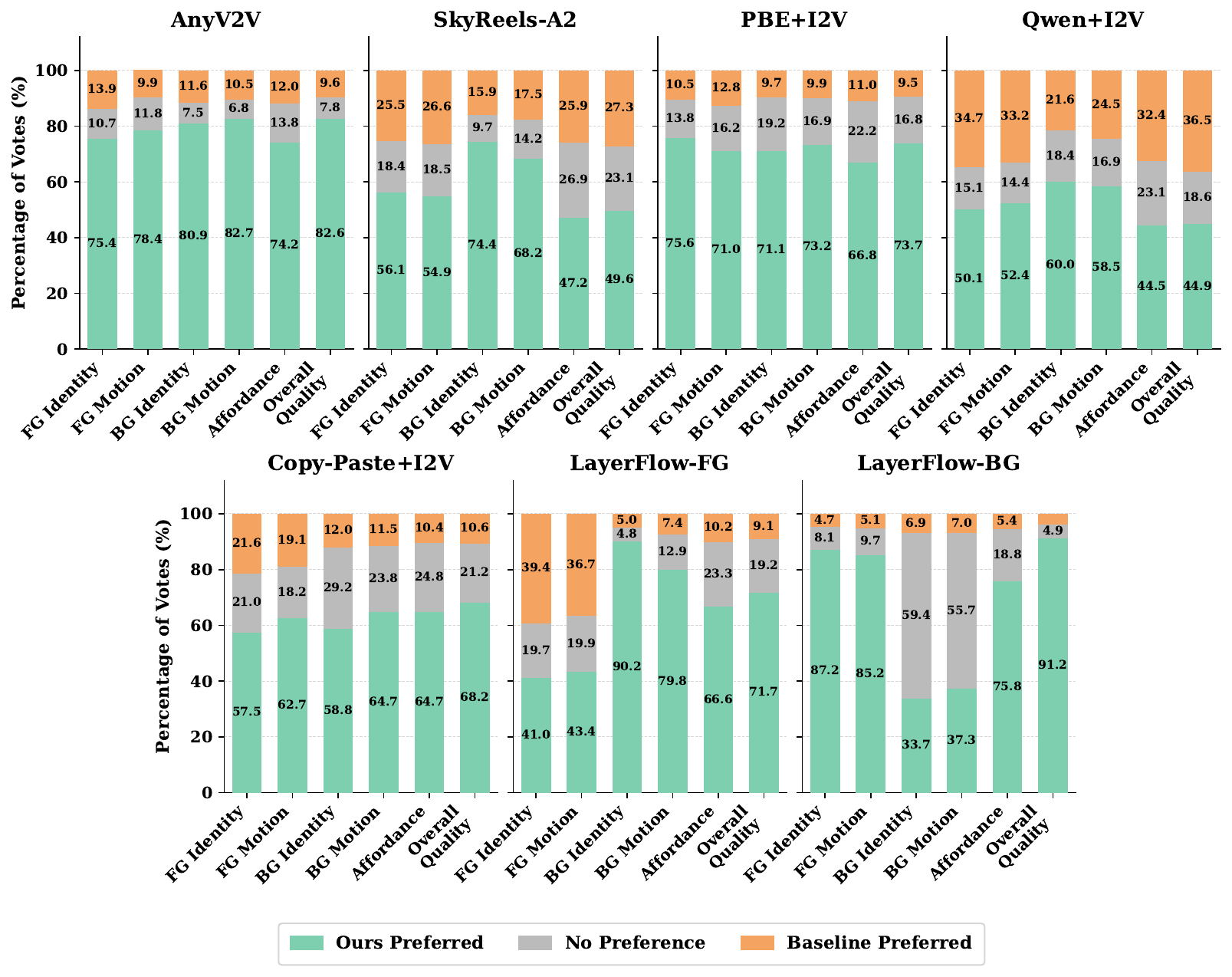}
    \caption{\textbf{\textit{User Study Results.}} Pairwise preference win/tie/lose rates comparing our method (\textbf{StM}) against five baselines across six criteria. Preference for StM is shown in Green, ties in Grey, and preference for the baseline in Orange. The results demonstrate that StM consistently outperforms all baselines, achieving its highest preference rates in motion and identity preservation.}
    \label{fig:user_study_results}
    \includegraphics[width=0.7\textwidth]{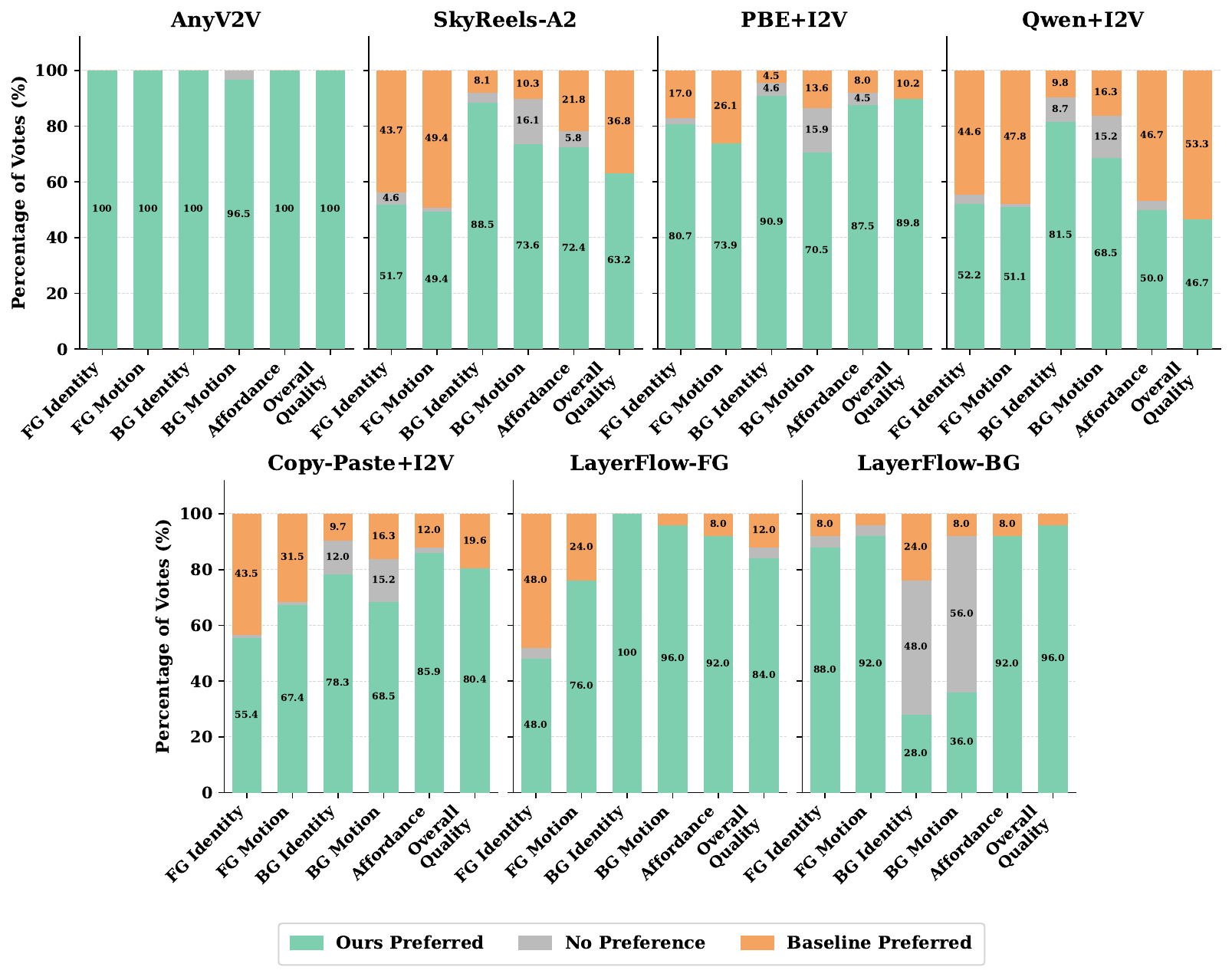}
    \caption{\textbf{\textit{VLLM-as-a-Judge Results.}} Results from our automated evaluation using Gemini 2.5 Pro, which follows the exact pairwise comparison protocol employed in the human study. The VLLM judge demonstrates a strong alignment with human preferences, consistently favoring our \textbf{StM} in key motion and identity metrics.}
    \label{fig:vllm_results}
\end{figure*}
\subsection{Test Benchmark}
\label{sec:test_benchmark}
\begin{table}[t!]
    \centering
    \caption{Comparison of recent baseline methods by task, venue, and evaluation size. Our proposed method, StM, focuses on video composition.}
    \label{tab:baseline_comparison}
    \begin{tabular}{@{}lllc@{}}
        \toprule
        \textbf{Method} & \textbf{Task} & \textbf{Venue} & \textbf{Size} \\ \midrule
        SkyReels & Img Comp. & Arxiv'25 & 50 \\
        LayerFlow & Txt2Layer & SIG'25 & 50 \\
        AnyV2V & Vid Edit & TMLR'25 & 89 \\
        \textbf{StM (Ours)} & \textbf{Vid Comp.} & \textbf{N/A} & \textbf{93} \\ \bottomrule
    \end{tabular}
\end{table}
For the final evaluation, we curated a dedicated test benchmark of 93 unique, unseen triplets (foreground video, background video, text prompt) from a held-out set. Crucially, the foreground and background components for each test sample were intentionally sourced from different original videos. This rigorous design robustly measures the model's ability to generalize and synthesize plausible, novel interactions between disparate visual elements. The source videos for this benchmark were primarily selected from the DAVIS dataset~\cite{perazzi2016benchmark}, with additional samples sourced from the Animal Kingdom dataset~\cite{ng2022animal}. As shown in Table~\ref{tab:baseline_comparison}, our evaluation dataset size is within the range of baseline models, covering diverse backgrounds, foregrounds, and lighting conditions.
\subsection{Decomposer-Independent Evaluation}
\label{sec:decomposer_independent}
Since all reported metrics depend on a Decomposer to extract layers from the generated videos, the choice of segmenter could, in principle, bias the comparison. To rule out such bias, we re-evaluate \emph{every} method using three independent decomposition pipelines: our motion-aware segmenter used in the main paper (Seg-Motion), SAM-2 + Grounding-DINO (which uses no motion prior), and SAM-3 (Table~\ref{tab:eval_ablation}). While the absolute scores shift with the choice of Decomposer, StM maintains its lead on \emph{every} metric across \emph{all three} pipelines. M5 is Decomposer-free and is therefore omitted from this comparison. We note that no established method exists for this exact task; the diverse set of baselines—including video-based methods (LayerFlow-FG/BG, AnyV2V) with full FG motion access—is provided to ensure a comprehensive comparison, and StM consistently outperforms all of them.
\begin{table*}[t!]
\centering
\setlength{\tabcolsep}{3pt}
\renewcommand{\arraystretch}{0.9}
\caption{\textbf{Stability under different Decomposers.} All methods are evaluated with three independent decomposition pipelines. StM leads on every metric (M1--M4) across all three.}
\label{tab:eval_ablation}
\resizebox{\textwidth}{!}{%
\begin{tabular}{l|c|c|c}
\toprule
                          & Seg-Motion (paper)               & SAM2 + Grounding-DINO            & SAM-3                            \\
                          & M1$\uparrow$ / M2$\uparrow$ / M3$\downarrow$ / M4$\downarrow$
                          & M1$\uparrow$ / M2$\uparrow$ / M3$\downarrow$ / M4$\downarrow$
                          & M1$\uparrow$ / M2$\uparrow$ / M3$\downarrow$ / M4$\downarrow$ \\
\midrule
Copy-Paste + I2V          & 83.1 / 85.0 / 1.6 / 184.6    & 86.5 / 89.2 / 1.0 / 251.3    & 82.4 / 85.2 / 1.4 / 258.1    \\
PBE + I2V                 & 73.5 / 80.2 / 2.5 /  98.1    & 81.1 / 86.5 / 1.7 /  92.9    & 77.5 / 84.2 / 1.6 /  95.5    \\
Qwen + I2V                & 82.0 / 72.4 / 1.7 /  74.8    & 84.5 / 75.6 / 1.0 /  95.8    & 83.8 / 75.6 / 1.1 /  95.1    \\
SkyReels                  & 80.2 / 75.2 / 1.8 / 279.2    & 81.3 / 76.9 / 1.2 / 381.1    & 79.7 / 74.8 / 1.1 / 362.1    \\
AnyV2V                    & 77.7 / 56.4 / 1.7 / 155.0    & 76.3 / 56.7 / 1.0 / 165.3    & 75.9 / 56.6 / 1.4 / 175.4    \\
LayerFlow-FG              & 83.0 / 60.2 / 1.5 / 143.5    & 59.3 / 50.7 / 5.9 / 186.8    & 54.0 / 26.4 / 4.0 / 176.2    \\
LayerFlow-BG              & 50.3 / 91.7 / 2.4 /  24.7    & 58.7 / 83.2 / 5.7 /  53.6    & 51.9 / 80.2 / 4.1 /  79.1    \\
\midrule
\textbf{StM\,(ours)}      & {\bf 84.8} / {\bf 92.9} / {\bf 1.2} / {\bf 16.4}
                          & {\bf 86.9} / {\bf 93.4} / {\bf 0.8} / {\bf 17.6}
                          & {\bf 83.9} / {\bf 90.6} / {\bf 1.1} / {\bf 16.9} \\
\bottomrule
\end{tabular}}
\end{table*}
\section{Additional Qualitative Results and Instructions}
\label{sec:qualitative_details}
\subsection{User Study Details}
\label{sec:user_study}
\begin{figure*}[t]
    \centering
    \begin{subfigure}[b]{0.48\textwidth}
        \includegraphics[width=\textwidth]{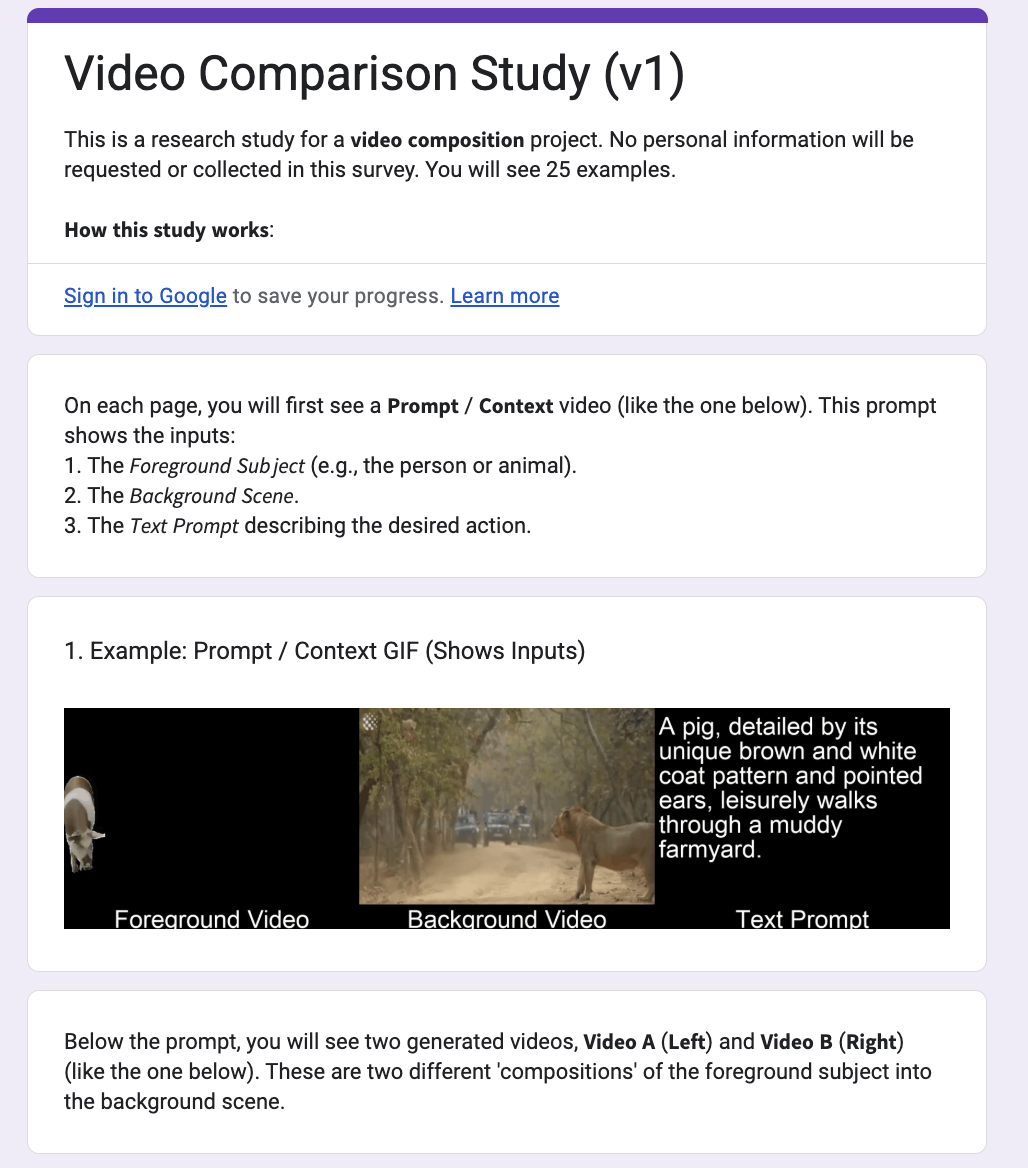}
        \caption{Instructions: Study Overview \& Inputs}
        \label{fig:us1}
    \end{subfigure}
    \hfill
    \begin{subfigure}[b]{0.48\textwidth}
        \includegraphics[width=\textwidth]{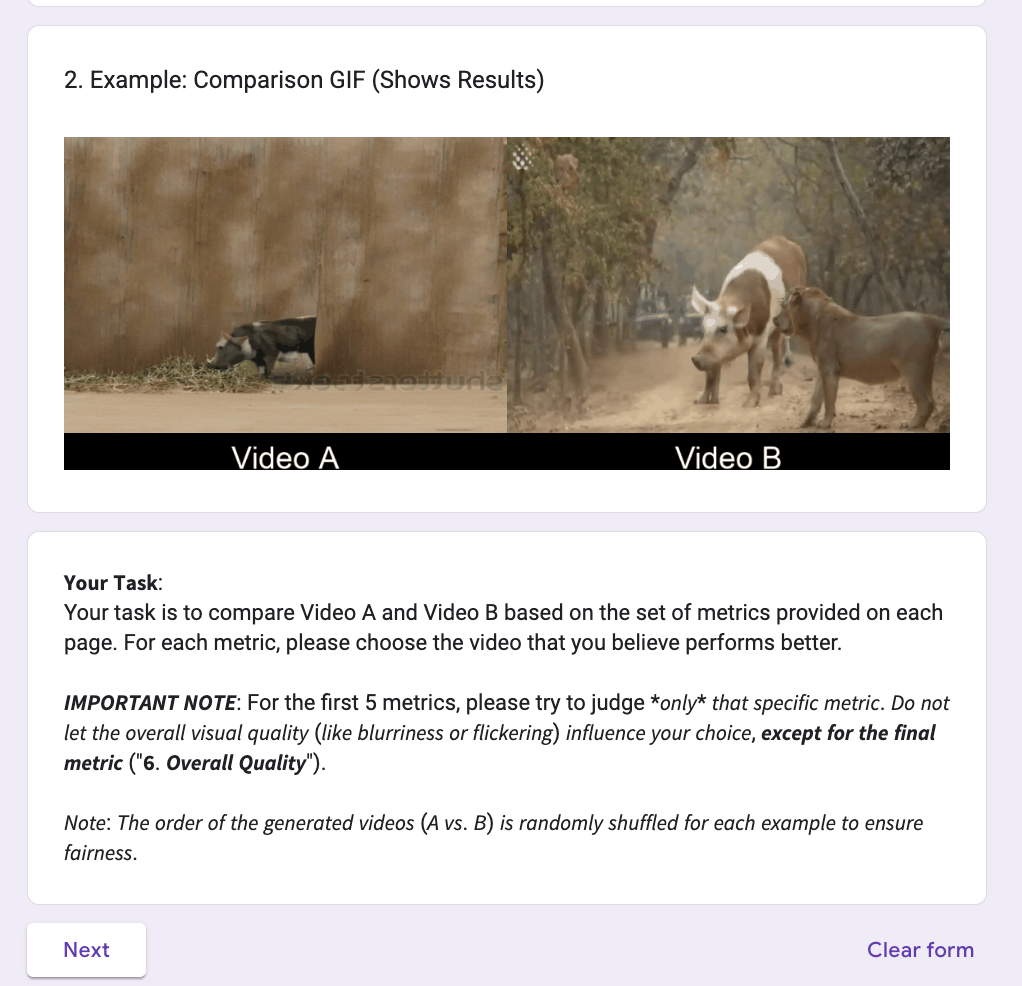}
        \caption{Instructions: Example Comparison}
        \label{fig:us2}
    \end{subfigure}
    \vspace{0.5em}
    \begin{subfigure}[b]{0.48\textwidth}
        \includegraphics[width=\textwidth]{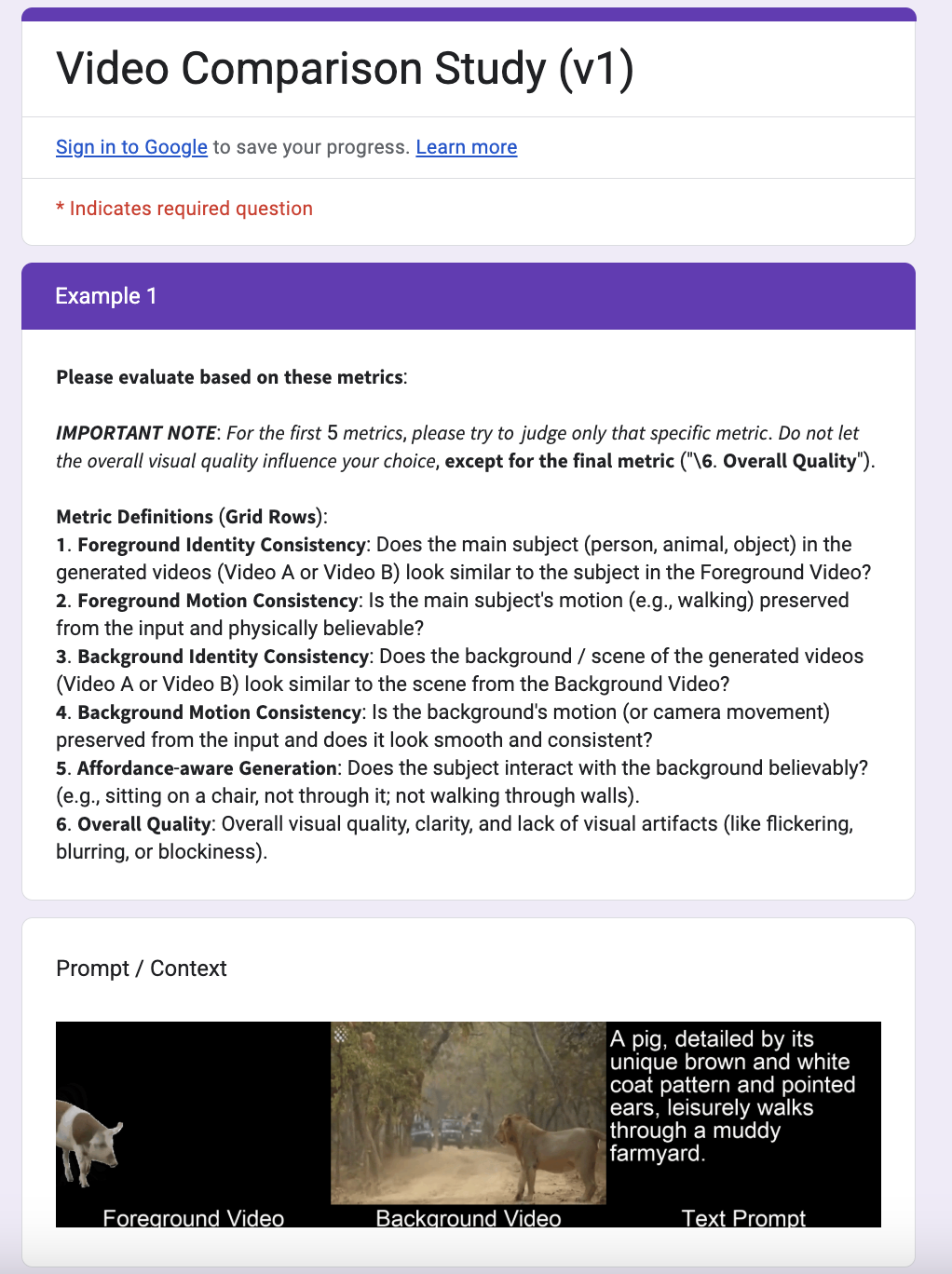}
        \caption{Question Interface: Context \& Metrics}
        \label{fig:us3}
    \end{subfigure}
    \hfill
    \begin{subfigure}[b]{0.48\textwidth}
        \includegraphics[width=\textwidth]{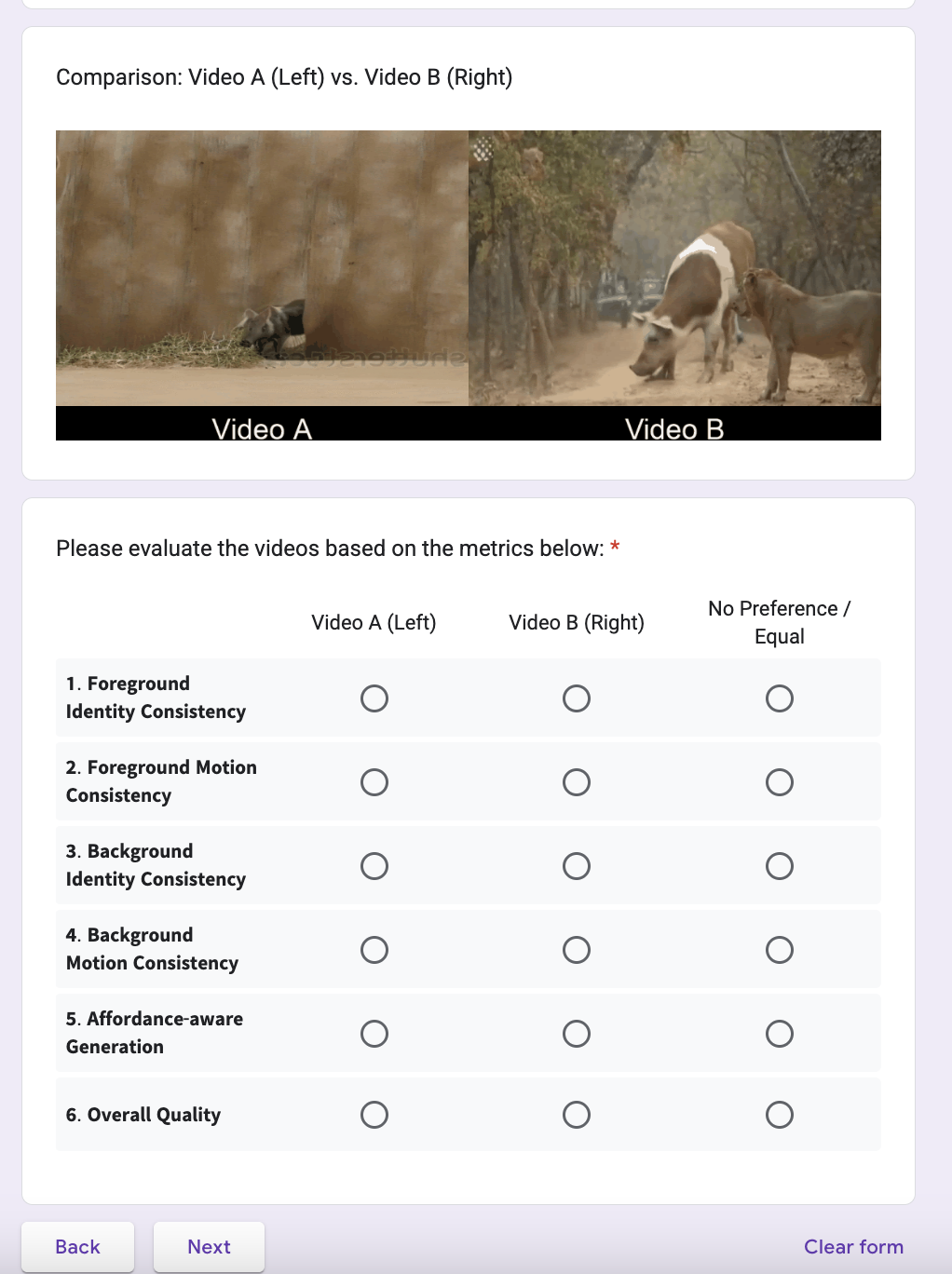}
        \caption{Question Interface: Video Comparison \& Rating}
        \label{fig:us4}
    \end{subfigure}
    \caption{\textbf{User Study Interface.} (a)-(b) The initial instruction pages presented to the subject, defining the inputs and demonstrating the task. (c)-(d) An example of a single question page. This interface—showing the context, metric definitions, and side-by-side video comparison—is repeated 25 times, once for each randomly sampled test case in the study.}
    \label{fig:user_study_interface}
\end{figure*}

Our pairwise preference study involved 50 subjects recruited via the Prolific platform. The evaluators performed a total of 25 pairwise comparisons, randomly sampled from our test dataset, and were asked to select the superior output video (StM vs. a baseline) based on five distinct criteria.
The user study procedure, illustrated by the interface layout in Figure~\ref{fig:user_study_interface}, began with an instructional phase (Figures~\ref{fig:us1} and \ref{fig:us2}). This phase explained the required task inputs and provided an example comparison. Evaluators then proceeded to the evaluation phase, which consisted of a series of 25 questions. As shown in Figures~\ref{fig:us3} and \ref{fig:us4}, each question page displayed the reference context, metric definitions, and the side-by-side video comparison for rating. To mitigate bias, the left-right presentation order of the compared methods was randomly shuffled for each question. This question format is repeated for all 25 samples for every baseline comparison.
The criteria used were:
\begin{itemize}
    \item \textbf{Identity Preservation (FG \& BG):} Measures visual identity fidelity of the foreground subject and background scene.
    \item \textbf{Motion Alignment (FG \& BG):} Assesses the fidelity of subject and camera/scene motion.
    \item \textbf{FG-BG Harmony:} Assesses plausible interaction and affordance-awareness.
    \item \textbf{Overall Quality:} A holistic judgment of realism, artifacts, and coherence.
\end{itemize}
\subsection{Affordance-Aware Evaluation}
\label{sec:affordance_eval}
Because no dedicated automated metric exists for affordance, standard practice pairs the motion metrics (M3, M4) with VLM and user judgments. Our core evaluation already covers this through M3/M4 and the FG-BG Harmony criterion, where StM wins by 54.95--97.96\%. For finer granularity, we extend both the VLLM-as-a-Judge and user-study protocols with six dedicated affordance criteria, comparing StM against video baselines (AnyV2V, LayerFlow-FG/BG) and an image baseline (PBE+I2V). The six criteria are: \emph{Physical Interaction} (ground contact, water ripples at contact, weight-bearing); \emph{Scale} (subject size believable relative to the scene); \emph{Placement} (subject placed in affordance-consistent locations); \emph{Action--Scene Affordance} (the prompted action matches the scene's capability); \emph{Lighting \& Shadow Consistency} (shading and shadow direction/color/hardness match the scene); and \emph{Occlusion / Depth Ordering} (the subject correctly goes behind or in front of scene objects rather than being pasted on top). As reported in Table~\ref{tab:phys}, StM wins all VLM and user judgments (66.7--97.3\%).
\begin{table*}[t!]
    \centering
    \setlength{\tabcolsep}{3pt}
    \renewcommand{\arraystretch}{0.9}
    \caption{\textbf{Affordance pairwise} (StM\ \w{win rate}, \%, VLM~/~User). All cells show StM\ winning ($\geq$50\%).}
    \label{tab:phys}
    \resizebox{\textwidth}{!}{%
    \begin{tabular}{lcccccc}
    \toprule
    vs.\ Baseline & Phys.\ Inter. & Scale & Placement & Act.\ Scene & Light./Shadow & Occl.\ Depth \\
    \midrule
    PBE + I2V       & \w{87.0} / \w{75.9} & \w{77.2} / \w{73.1} & \w{91.3} / \w{75.1} & \w{79.3} / \w{77.8} & \w{93.5} / \w{79.1} & \w{92.3} / \w{77.3} \\
    AnyV2V          & \w{91.5} / \w{78.4} & \w{76.8} / \w{74.4} & \w{86.6} / \w{76.6} & \w{90.2} / \w{75.2} & \w{93.9} / \w{81.5} & \w{90.1} / \w{78.9} \\
    LF-FG           & \w{78.9} / \w{81.5} & \w{66.7} / \w{77.6} & \w{81.1} / \w{79.4} & \w{71.7} / \w{81.6} & \w{86.0} / \w{79.9} & \w{88.6} / \w{80.2} \\
    LF-BG           & \w{94.4} / \w{97.3} & \w{85.7} / \w{92.4} & \w{92.4} / \w{95.1} & \w{90.1} / \w{96.9} & \w{90.2} / \w{96.3} & \w{90.2} / \w{96.2} \\
    \bottomrule
    \end{tabular}}
\end{table*}
\subsection{VLLM-as-a-Judge Instructions}
\label{sec:vllm_judge}
We utilized \textbf{Gemini 2.5 Pro} (Public version) as a Vision-Language Large Model (VLLM) to act as an automated judge on the full test dataset. The VLLM was provided with the same pairwise comparison task as the human subjects. To ensure fair evaluation, we employed the specific system instructions listed below for each metric. The model was instructed to output a single character: A, B, or N (No Preference).

\noindent\textbf{a. Foreground Identity Consistency:}
\begin{quote}
\textit{"You are a video analysis tool. You will be given a Reference Foreground video and two generated videos (Video A, Video B). IMPORTANT: Judge *only* this specific metric. Do not let overall visual quality influence your choice. Metric: Foreground Identity Consistency. Question: Which generated video (Video A or Video B) better preserves the appearance of the subject (person, animal, object) from the Reference Foreground video?"}
\end{quote}
\noindent\textbf{b. Foreground Motion Consistency:}
\begin{quote}
\textit{"You are a video analysis tool. You will be given a Reference Foreground video and two generated videos (Video A, Video B). IMPORTANT: Judge *only* this specific metric. Do not let overall visual quality influence your choice. Metric: Foreground Motion Consistency. Question: Which generated video (Video A or Video B) better preserves the subject's motion from the Reference Foreground video and looks more physically believable?"}
\end{quote}
\noindent\textbf{c. Background Identity Consistency:}
\begin{quote}
\textit{"You are a video analysis tool. You will be given a Reference Background video and two generated videos (Video A, Video B). IMPORTANT: Judge *only* this specific metric. Do not let overall visual quality influence your choice. Metric: Background Identity Consistency. Question: Which generated video (Video A or Video B) better preserves the appearance of the background scene from the Reference Background video?"}
\end{quote}
\noindent\textbf{d. Background Motion Consistency:}
\begin{quote}
\textit{"You are a video analysis tool. You will be given a Reference Background video and two generated videos (Video A, Video B). IMPORTANT: Judge *only* this specific metric. Do not let overall visual quality influence your choice. Metric: Background Motion Consistency. Question: Which generated video (Video A or Video B) better preserves the background's motion (or camera movement) from the Reference Background video, and which looks smoother?"}
\end{quote}
\noindent\textbf{e. Affordance-aware Generation:}
\begin{quote}
\textit{"You are a video analysis tool. You will be given a Reference Foreground, a Reference Background, and two generated videos (Video A, Video B). IMPORTANT: Judge *only* this specific metric. Do not let overall visual quality influence your choice. Metric: Affordance-aware Generation. Question: Which generated video (Video A or Video B) shows a more believable interaction between the subject and the background? (e.g., sitting *on* a chair, not *through* it; not walking through walls)."}
\end{quote}
\noindent\textbf{f. Overall Quality:}
\begin{quote}
\textit{"You are a video analysis tool. You will be given a Reference Foreground, a Reference Background, and two generated videos (Video A, Video B). Metric: Overall Quality. Question: Which generated video (Video A or Video B) has the best overall visual quality, clarity, and fewest visual artifacts (like flickering, blurring, or blockiness)?"}
\end{quote}

\subsection{VLLM-as-a-Judge Affordance Instructions}
\label{sec:vllm_judge_affordance}
For the finer-grained affordance evaluation (Table~\ref{tab:phys}), we employed the same pairwise protocol with six dedicated affordance criteria. As before, the VLLM was instructed to output a single character: A, B, or N (No Preference).

\noindent\textbf{a. Physical Interaction:}
\begin{quote}
\textit{"You are a video analysis tool. You will be given a Reference Foreground, a Reference Background, and two generated videos (Video A, Video B). IMPORTANT: Judge *only* this specific metric. Do not let overall visual quality influence your choice. Metric: Physical Interaction. Question: Which generated video (Video A or Video B) shows more physically plausible contact between the subject and the scene (e.g., feet making firm ground contact and bearing weight, water rippling or splashing where the subject touches it), rather than the subject hovering, sinking, or sliding unnaturally?"}
\end{quote}

\noindent\textbf{b. Scale:}
\begin{quote}
\textit{"You are a video analysis tool. You will be given a Reference Foreground, a Reference Background, and two generated videos (Video A, Video B). IMPORTANT: Judge *only* this specific metric. Do not let overall visual quality influence your choice. Metric: Scale. Question: Which generated video (Video A or Video B) renders the subject at a more believable size relative to the surrounding scene and its objects, avoiding a subject that is implausibly large or small for the environment?"}
\end{quote}

\noindent\textbf{c. Placement:}
\begin{quote}
\textit{"You are a video analysis tool. You will be given a Reference Foreground, a Reference Background, and two generated videos (Video A, Video B). IMPORTANT: Judge *only* this specific metric. Do not let overall visual quality influence your choice. Metric: Placement. Question: Which generated video (Video A or Video B) positions the subject in a location that is consistent with where it could plausibly exist in the scene (e.g., a boat on water rather than on a road, an animal on the ground rather than floating in the air)?"}
\end{quote}

\noindent\textbf{d. Action--Scene Affordance:}
\begin{quote}
\textit{"You are a video analysis tool. You will be given a Reference Foreground, a Reference Background, and two generated videos (Video A, Video B). IMPORTANT: Judge *only* this specific metric. Do not let overall visual quality influence your choice. Metric: Action--Scene Affordance. Question: Which generated video (Video A or Video B) better matches the subject's action to what the scene actually affords (e.g., swimming where there is water, walking where there is a walkable surface), so that the action is consistent with the environment rather than physically impossible within it?"}
\end{quote}

\noindent\textbf{e. Lighting \& Shadow Consistency:}
\begin{quote}
\textit{"You are a video analysis tool. You will be given a Reference Foreground, a Reference Background, and two generated videos (Video A, Video B). IMPORTANT: Judge *only* this specific metric. Do not let overall visual quality influence your choice. Metric: Lighting \& Shadow Consistency. Question: Which generated video (Video A or Video B) better harmonizes the subject's shading and shadows with the scene, such that shadow direction, color, and hardness, as well as the lighting on the subject, are consistent with the background's light sources?"}
\end{quote}

\noindent\textbf{f. Occlusion / Depth Ordering:}
\begin{quote}
\textit{"You are a video analysis tool. You will be given a Reference Foreground, a Reference Background, and two generated videos (Video A, Video B). IMPORTANT: Judge *only* this specific metric. Do not let overall visual quality influence your choice. Metric: Occlusion / Depth Ordering. Question: Which generated video (Video A or Video B) better respects depth ordering, with the subject correctly passing behind or in front of scene objects according to their relative depth, rather than appearing pasted on top of everything?"}
\end{quote}
\section{Computational Cost}
\label{sec:efficiency}
The computational overhead of the \MethodName\ framework is concentrated in the \textit{\textbf{training phase}}, because the transformation-aware training pipeline encodes three distinct video inputs ($V_{org}$, $V_{bg}$, $\tilde{V}_{fg}$) through the frozen Space-Time VAE. The Composer is fine-tuned for 20K iterations on 16 NVIDIA H100 GPUs with a total batch size of 64.

At \textit{\textbf{inference}} time, the process is efficient. We strictly preserve the architecture of the base model (CogVideoX-I2V~\cite{yang2025cogvideox}), adding only a lightweight projection layer at the input stage. Since this projection is negligible in size relative to the transformer backbone, our method incurs no additional computational complexity during the iterative denoising process. Consequently, the inference latency is effectively identical to the base I2V model, with the only overhead being the one-time encoding of the additional video layers.

\end{document}